\documentclass[journal,twoside,web]{IEEEtran}

\usepackage{verbatim}
\usepackage[binary-units=true]{siunitx}
\usepackage{times}
\usepackage{amsmath,amssymb}
\usepackage{accents}
\usepackage{graphicx}
\usepackage{bm}
\let\labelindent\relax
\usepackage[inline]{enumitem}
\usepackage{array}
\usepackage{multicol}
\usepackage{multirow}
\usepackage{tabulary}
\usepackage{booktabs}
\usepackage{subfig}
\usepackage{todonotes}
\usepackage{cite}
\usepackage[hidelinks]{hyperref}
\usepackage{algorithm}% http://ctan.org/pkg/algorithm
\usepackage{algpseudocode}% http://ctan.org/pkg/algorithmicx
\usepackage{breqn}
\usepackage{tcolorbox}
\usepackage{mathtools}
\usepackage{tikz}
\usetikzlibrary{arrows}
\usetikzlibrary{calc}
\usetikzlibrary{arrows,decorations.markings}
\usepackage{balance}

\allowdisplaybreaks

\renewcommand{\exp}[1]{\mathrm{e}^{#1}}             %Exponential
\DeclareMathOperator{\diag}{diag}

\DeclareMathOperator*{\mini}{min.}

\definecolor{light-gray}{gray}{0.95}
\definecolor{dark-gray}{gray}{0.5}
\definecolor{mygray}{gray}{0.75}
\newcommand{\BIN}{\begin{bmatrix}}
\newcommand{\BOUT}{\end{bmatrix}}

\newcommand{\ue}{\underline{e}}
\newcommand{\uw}{\underline{w}}

\newcommand{\uJ}{\underline{J}}

\newcommand{\ulambda}{\underline{\lambda}}

\pgfdeclarelayer{bg1}   
\pgfdeclarelayer{bg}  
\pgfsetlayers{bg1,bg,main}  

\definecolor{orange}{rgb}{0.99,0.69,0.07}
\definecolor{lightgray}{gray}{0.85}
\definecolor{light-gray}{gray}{0.95}
\definecolor{dark-gray}{gray}{0.5}

\tikzset{cross/.style={cross out, draw=black, minimum size=2*(#1-\pgflinewidth), inner sep=0pt, outer sep=0pt},
cross/.default={1pt}}

 \makeatletter
 \newcommand\fs@spaceruled{\def\@fs@cfont{\bfseries}\let\@fs@capt\floatc@ruled
   \def\@fs@pre{\vspace{5pt}\hrule height.8pt depth0pt \kern2pt}%
   \def\@fs@post{\kern2pt\hrule\relax}%
   \def\@fs@mid{\kern2pt\hrule\kern2pt}%
   \let\@fs@iftopcapt\iftrue}
 \makeatother

\usepackage{cite}
\usepackage{amsmath,amssymb,amsfonts}
\usepackage{graphicx}
\usepackage{textcomp}

\usepackage{nomencl}
\makenomenclature

\begin{document}
\title{The hierarchical Newton's method for numerically stable prioritized dynamic control}
\author{Kai Pfeiffer, Adrien Escande, Pierre Gergondet, Abderrahmane Kheddar, \IEEEmembership{Fellow, IEEE}
\thanks{This research was supported in part by the CNRS-AIST-AIRBUS Joint Research Program, and the EU H2020 COMANOID project.}
\thanks{K.~Pfeiffer was with the CNRS-AIST Joint Robotics Laboratory (JRL), IRL, Tsukuba, Japan. He is now with the Nanyang Technological University, Singapore.}%
	\thanks{A. Escande, P. Gergondet and A. Kheddar are with the CNRS-AIST  Joint  Robotics  Laboratory  (JRL), UMI3218/RL, Tsukuba, Japan.}%
	\thanks{A. Kheddar is also with the CNRS-University of Montpellier, LIRMM, UMR5506, Interactive Digital Human, France.)}%
}

\maketitle

	\begin{abstract}
		This work links optimization approaches from hierarchical least-squares programming to instantaneous prioritized whole-body robot control. Concretely, we formulate the hierarchical Newton's method which solves prioritized non-linear least-squares problems in a numerically stable fashion even in the presence of kinematic and algorithmic singularities of the approximated kinematic constraints. These results are then transferred to control problems which exhibit the additional variability of time. This is necessary in order to formulate acceleration based controllers and to incorporate the second order dynamics. However, we show that the Newton's method without complicated adaptations is not appropriate in the acceleration domain. We therefore formulate a velocity based controller which exhibits second order proportional derivative convergence characteristics. Our developments are verified in toy robot control scenarios as well as in complex robot experiments which stress the importance of prioritized control and its singularity resolution.	
	\end{abstract}
	

	\section{Introduction}

%	\cite{Johansen2004,}
	\IEEEPARstart{C}{ontrol} hierarchies formulated as optimization problems can be solved very efficiently, see e.g.,~\cite{Escande2014}.
	However, the evaluation of this solver was confined to simulations or well-calibrated experiments due to the lack of appropriate methods to resolve singularities in hierarchies. If unresolved, they lead to unpredictable instabilities and risky behaviors~\cite{nakamura1986}. With this paper, and in the vein of~\cite{Escande2014}, we aim at solving this drawback from an optimization based perspective. 
	
	Hierarchical inverse dynamics problems have been approached for the longest time by projector based methods~\cite{Siciliano1991}. While they provide a very intuitive insight into the physics of the problem, they lack the powerful tools of optimization. This stretches from the incorporation of inequality constraints to fast solver formulations. While some extensions for example for inequality constraints have been proposed~\cite{Chaumette2000,Mansard2005}, only with the introduction of optimization based approaches such problems could be overcome in a holistic fashion~\cite{abe:sca:2007,collette:humanoids:2007}. Nowadays, optimization based control is widespread in robotics and has been applied in numerous works~\cite{feng:humanoids:2013,kuindersma:icra:2014,vaillant:auro:2016,pfeiffer2017}. Especially the implementation of quadratic programming (QP) based solvers acted as a catalyst for this process. In classical constrained QP based control approaches, control tasks can be specified in a two-level hierarchy  where the equation of motion and all the corresponding dynamics constraints are put on the constraints level, while the robot control is put on the objective level, see e.g.~\cite{Bouyarmane2019}. With the rise of new hierarchical solvers~\cite{Kanoun2011,Escande2014,Herzog2016} the robot control can handle more priority levels, for example respecting joint limits over the control of any other task. This allows designing very safe controllers, strictly prioritizing safety, physical stability constraints and objective tasks. Switches in the hierarchical ordering of tasks can be smoothly achieved as for example proposed in~\cite{Arechavaleta2017}.  
	
	Problems arise if tasks on different levels of the hierarchy get into conflict. Such algorithmic singularities need to be resolved alongside kinematic singularities. Otherwise the near rank deficiency of the Jacobian of the task linearizations --or its projection onto the Jacobians of higher priority tasks in case of algorithmic singularities-- leads to numerical instability with high joint velocities. Kinematic singularities can be quantified with the manipulability measure~\cite{Yoshikawa1985} which can be maximized for singularity avoidance in an optimization setting~\cite{Johansen2004}. Another measure is the singularity index~\cite{Leeghim2009} which has been proposed in the context of gimbal attitude control of spacecraft. Analytical robot workspace analysis~\cite{Maciejewski1988,Khalil2002,Bianco2020} enables the prediction and avoidance of kinematic singularities. In contrast, algorithmic singularities depend on the conflict with higher priority tasks at the current robot configuration and therefore are harder to analyze~\cite{Chiaverini1997}. Furthermore, hierarchical solvers are intended to be employed on humanoid robots in any industrial setting, e.g. aircraft manufacturing~\cite{kheddar2019ram} or construction~\cite{kumagai2019ram}. Here end-users are requested to specify a set of usual tasks (set-point, tip force control, trajectory tracking...)~\cite{abe:sca:2007,vaillant:auro:2016} under various predefined constraints (joint limits...) or constraints updated from online sensor readings especially when concerned with the physical stability of the robot~\cite{Nishiwaki2009,Mason2016,bonnet:hal-02048085}. Potential issues might arise if for example sensor readings give targets that are outside of the feasible workspace of the robot. Here engineers without deeper understanding of possible task conflicts and the resulting singularities should be able to set up robot problems safely, quickly and easily. 	
	
	It is in the vein of optimization that singularity resolution methods like `damping' in the projector based approach~\cite{nakamura1986,ChiaveriniSiciliano1994,Buss2005,Sugihara2011,Harish2016} can be interpreted as their equivalent in optimization, here the Levenberg-Marquardt (LM) algorithm. At the same time, the LM algorithm can be interpreted as approximating the second order derivatives of the Taylor expansion of the quadratic task error norm~\cite{Dennis1981,Deo:1993} as a weighted identity matrix.
	In our previous work~\cite{Pfeiffer2018} we showed that using an approximation of the true second order derivatives by the  Broyden, Fletcher, Goldfarb and Shanno (BFGS) method~\cite{Broyden1970} yields better convergence. Additionally, the tuning of the damping parameter is not straightforward~\cite{Nenchev2000}.
	
	The previously presented Quasi-Newton method~\cite{Pfeiffer2018} was only aimed at resolving singularities in hierarchical kinematics based control problems. While kinematic control of robots can be sufficient for fixed base robots \cite{Wang2010}, this does not necessarily hold for legged humanoids with an un-actuated free-flyer base and unilateral friction contacts. Only with a model of the forces and torques acting on the robot's body the robot can aim to maintain a physically stable posture~\cite{Siciliano2007}. Therefore, we present ways to include the equation of motion and dynamics constraints into our scheme to generate physically feasible motions while borrowing approaches from numerical optimization.
	
\begin{figure*}[htp!]
\centering
\begin{tikzpicture}[line cap=rect]

\node[align=left,text width=5.4cm] (NLHLSP) at (-1,-1) {
\begin{tcolorbox}[title=Non-linear Hierarchical \\Least-Squares Program\\ (NL-HLSP)~\eqref{eq:NL-HLSP},boxsep=1pt,left=5pt,right=3pt,top=2pt,bottom=1pt] 
\begin{enumerate}
\item[L.$1$] $e_{1}^i(q,\dot{q},\ddot{q},\tau,\gamma)-w_{1}^i \leqq~0$\\
{\centering$\vdots$}
\item[L.$p$] $e_{p}^i(q,\dot{q},\ddot{q},\tau,\gamma)-w_{p}^i \leqq~0$
\end{enumerate}	
\end{tcolorbox}
};

\node[draw,align=left,text width=5cm] (Lin) at (4.85,-1) {
	\textit{Linearization at time $t_k$}\\
    \begin{itemize}[leftmargin=3mm]
    \item FK
    \begin{itemize}
    \item \textbf{\color{teal}\emph{Hierarchical Newton's method \eqref{eq:hessianHier},~\eqref{eq:Nopt} of control \eqref{eq:Ncontrol},~\eqref{eq:PDvel},~\eqref{eq:intEqOfMot}}}
    \item Hierarchical Quasi-Newton method~\cite{Pfeiffer2018} \textbf{\color{teal}\emph{of control \eqref{eq:Ncontrol},~\eqref{eq:PDvel},~\eqref{eq:intEqOfMot}}}
    \item {\color{red}Hierachical Gauss-Newton algorithm of control}
    \end{itemize}
    \item FD
    \begin{itemize}
    \item Instantiation (keeping $q_k$, $\dot{q}_k$ const.)
    \end{itemize}
    \end{itemize}
};

\node[text width=6cm] (HLSP) at (11.,-1) {
	\begin{tcolorbox}[title=Hierarchical Least-Squares\\ Program (HLSP)~\eqref{eq:generalHierarchy},boxsep=1pt,left=5pt,right=3pt,top=2pt,bottom=1pt] 
	\begin{enumerate}
	\item[L.$0$] $\Vert\dot{q}_{k+1}\Vert_{\infty} \leq \rho$\\
	\item[L.$1$] $A_{k,1}^ix_k+b_{k,1}^i-w_{k,1}^i\leqq~0$\\
	 $\vdots$
	\item[L.$p$] $A_{k,p+1}^ix_k + b_{k,p+1}^i - w_{k,p+1}^i \leqq~0$
	\end{enumerate}	
	\end{tcolorbox}
};

\node[draw,text width=3.5cm] (sol) at (11,-6) {
	\textit{Hierarchical Least-squares solver}~\cite{Escande2014,dimitrov:2015}\\ 
	$x_k = \protect\begin{bmatrix}\dot{q}_{k+1}^T, \tau_k^T, \gamma_k^T\protect\end{bmatrix}^T$
};

\node[draw,text width=4cm] (int) at (4.75,-6) {
	\textit{Integration} \\
		$q_{k+1}$, $\dot{q}_{k+1}$, $\ddot{q}_{k+1}$, $\tau_{k}$, $\gamma_{k}$
};

\node[text width=5cm] (robot) at (-1,-6) { 
		{\includegraphics[width=0.65\columnwidth]{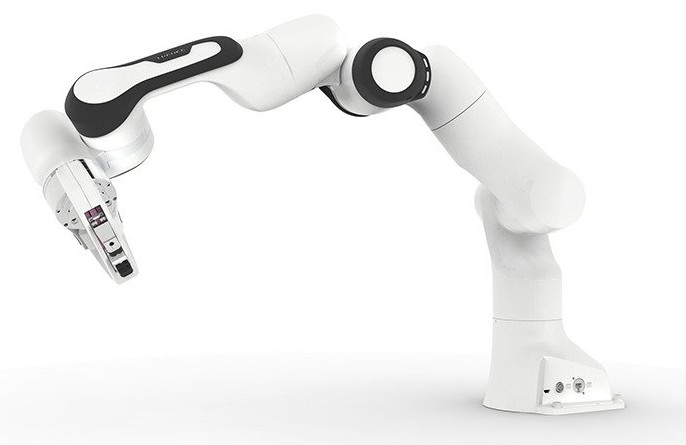}}
				{\includegraphics[width=0.3\columnwidth]{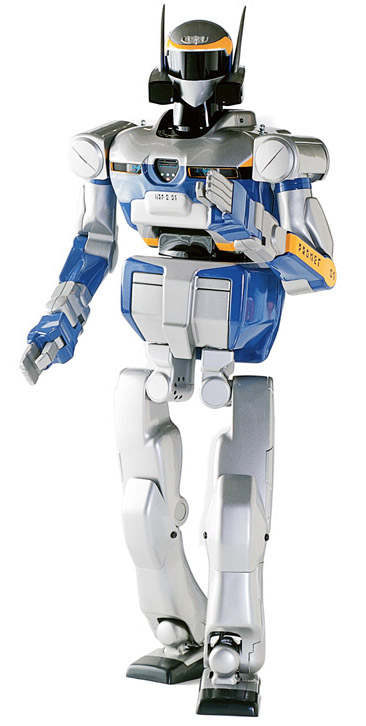}}
};

\coordinate (c1)   at ($(int.north) + (0,1.4)$);
\coordinate (c2)   at ($(int.north) + (-5.75,1.4)$);

\draw[decoration={markings,mark=at position 1 with
		{\arrow[scale=3,>=stealth]{>}}},postaction={decorate}] ($(NLHLSP.east) + (-0.15,0)$) -- (Lin.west);
\draw[decoration={markings,mark=at position 1 with
	{\arrow[scale=3,>=stealth]{>}}},postaction={decorate}] (Lin.east) -- ($(HLSP.west) + (0.15,0)$);
\draw[decoration={markings,mark=at position 1 with
	{\arrow[scale=3,>=stealth]{>}}},postaction={decorate}] ($(HLSP.south) + (0,0.125)$) -- (sol);
\draw[decoration={markings,mark=at position 1 with
	{\arrow[scale=3,>=stealth]{>}}},postaction={decorate}] (sol) -- (int);
\draw[decoration={markings,mark=at position 1 with
	{\arrow[scale=3,>=stealth]{>}}},postaction={decorate}] (int.north) -- (c1) -- (c2) -- ($(NLHLSP.south) + (0,0.125)$);
\draw[decoration={markings,mark=at position 1 with
	{\arrow[scale=3,>=stealth]{>}}},postaction={decorate}] (int) -- (robot);
\node[] (inc) at ($(int.north) + (1.5,0.5)$) {$t_{k+1}$ += $\Delta t$, $k$++};
\end{tikzpicture}
\caption{A symbolic overview of the sequential hierarchical least-squares programming with trust region (S-HLSP) to solve non-linear hierarchical least-squares programs (NL-HLSP)~\eqref{eq:NL-HLSP} with $p$ levels. This enables real-time robot control for a given set of non-linear forward dynamics (FD) or forward kinematics (FK) tasks. Contributions are marked in teal. Each priority level ($l$, L.) $1$~to~$p$ contains $m_l$ non-linear least-squares tasks (the norm notation is omitted for better readability). Each level~$l$ is constrained by the tasks of the previous priority levels $1$~to~$l-1$. First, the NL-HLSP is linearized either by optimization based methods (our contribution, the Hierarchical Newton's method of control~\eqref{eq:Ncontrol} for FK) or by instantiation (for FD). Linearization by the Hierarchical Gauss-Newton alogorithm of control leads to numerical instabilities in the case of kinematic and algorithmic singularities (red). The state $x_k$ is given by $x_k \coloneqq \protect \begin{bmatrix} \dot{q}_{k+1}^T&\Delta t{\tau}_k^T&\Delta t{\gamma}_k^T \protect \end{bmatrix}^T$~\eqref{eq:intEqOfMot}. The dependency on $q_k$, $\dot{q}_k$ of the linearization components $A_{k}(q_k,\dot{q}_k)$ and $b_k(q_k,\dot{q}_k)$ is omitted for better readability. The resulting (linear) hierarchical least-squares program (HLSP) (including a trust region constraint on $\dot{q}_{k+1}$, the number of priority levels is $p+1$) is then solved for $\dot{q}_{k+1}$, $\tau_k$ and $\gamma_k$ by a hierarchical active-set method. The integrated solution is sent to the robot. Finally, the next iteration of the S-HLSP is instantiated at the current state ${q}_{k}$, $\dot{q}_{k}$, $\ddot{q}_{k}$, $\tau_k$ and $\gamma_k$ after incrementation $k$++.}
%The images for the Franka Emika Panda and HRP-2 robots are taken from~\cite{panda} and~\cite{hrp2}, respectively.
\label{fig:diagram}
\end{figure*}
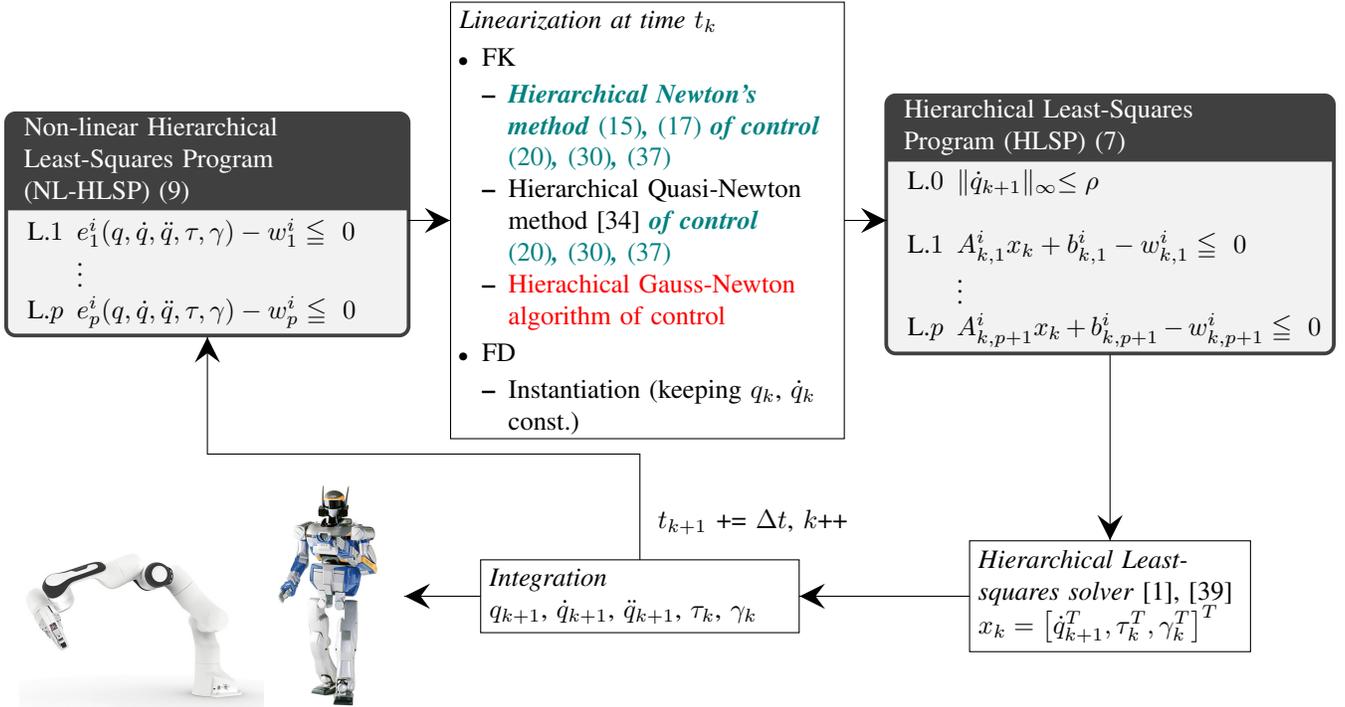

	This work then presents the following new contributions:
	\begin{itemize}
		\item We formulate the Newton's method for prioritized non-linear least-squares problems with a suitable formulation of the hierarchical Hessian (see sec.~\ref{sec:minNonLinGeoFct}).
		\item We adapt Newton's method, which is a tool from optimization, to constrained prioritized control (see sec.~\ref{sec:fromOptToCtrl});
		\item We show how regularization terms like damping negatively influence the exponential convergence of second-order motion controllers (see sec.~\ref{sec:dampAcc});
		\item Second-order motion controllers are suitably adapted to fit into the velocity based Newton's method of control (see sec.~\ref{sec:fromVelToAcc});
		\item The dynamics in form of the equation of motion are adapted accordingly and then integrated into the hierarchical control scheme (see sec.~\ref{sec:inclDyn});
		\item Experimental assessment of our developments with the HRP-2Kai humanoid robot in complex hierarchical control scenarios (see sec.~\ref{sec:validation}).
	\end{itemize}
A symbolic overview of this work is given in fig.~\ref{fig:diagram}. The corresponding nomenclature is summarized in	sec.~\ref{sec:nomenclature}.

	\section{Preliminaries}
We present hereafter the nomenclatures and variables naming that are used all along the paper.	
	\nomenclature[01]{\(n\)}{Dimension of joint configuration}
	\nomenclature[02]{\(n_f\)}{Number of free-flyer joints}
	\nomenclature[03]{\(n_c\)}{Number of contact points}
	\nomenclature[04]{\(m\)}{Dimension of task}
	
	\nomenclature[05]{\(t\)}{Continuous time}
	\nomenclature[06]{\(k\)}{Control iteration $k$}
	\nomenclature[07]{\(t_k\)}{Discrete time $t_k$ at iteration $k$}
	\nomenclature[08]{\(q(t)\in\mathbb{R}^n\)}{Kinematic configuration}
	\nomenclature[09]{\(f(q(t))\in\mathbb{R}^m\)}{Forward kinematics (FK)}
	\nomenclature[10]{\(f_d(t)\in\mathbb{R}^m\)}{Desired task value}
	\nomenclature[11]{\(e\in\mathbb{R}^m\)}{Task error; each dimension of the task may be weighted by a scalar}
	\nomenclature[12]{\(\leqq\)}{Equality or inequality task}
	\nomenclature[13]{\(J(q(t))\in\mathbb{R}^{m\times n}\)}{Jacobian of forward kinematics $f$}
	\nomenclature[14]{\({H}(q(t)\in\mathbb{R}^{m\times n\times n}\)}{Hessian of forward kinematics $f$}
	
	\nomenclature[15]{\(k_p\)}{Proportional gain}
	\nomenclature[16]{\(k_v\)}{Derivative gain}
	\nomenclature[17]{\({\dot{e}}^{\text{ctrl}}_{\text{P}}\in\mathbb{R}^m\)}{Velocity based P controller}
	\nomenclature[18]{\({\dot{e}}^{\text{ctrl}}_{\text{PD}}\in\mathbb{R}^m\)}{Velocity based PD controller}
	\nomenclature[19]{\({\ddot{e}}^{\text{ctrl}}_{\text{PD}}\in\mathbb{R}^m\)}{Acceleration based PD controller}
	
	\nomenclature[20]{\({M}({q_k})\in\mathbb{R}^{n\times n}\)}{Whole-body inertia matrix}
	\nomenclature[21]{\(N({q_k},{\dot{q}_k})\in\mathbb{R}^n\)}{Force vector}
	\nomenclature[22]{\(S\in\mathbb{R}^{n_f\times n}\)}{Selection matrix}
	\nomenclature[23]{\(\tau\in\mathbb{R}^n\)}{Joint torques}
	\nomenclature[24]{\(J_c\in\mathbb{R}^{n_c\times n}\)}{Contact Jacobian}	
	
	\nomenclature[25]{\(\Delta t\)}{Step in time $t$}
	\nomenclature[26]{\(\Delta q\in\mathbb{R}^n\)}{Step in configuration $q$}

	\nomenclature[27]{\(p\)}{Number of priority levels}
	\nomenclature[28]{\(l, L.\)}{Index of current priority level}
	\nomenclature[29]{\(w\in\mathbb{R}^m\)}{Slack variable}
	\nomenclature[30]{\(\lambda\in\mathbb{R}^m\)}{Lagrange multipliers}	
	\nomenclature[31]{\(\underline{a}_l\)}{Stacked vector
		$
		\begin{bmatrix}
		{v}_1^T & \cdots & {v}_l^T
		\end{bmatrix}^T
		$}
	\nomenclature[32]{\(\underline{A}_l\)}{Stacked matrix
		$
		\begin{bmatrix}
		{A}_1^T & \cdots & {A}_l^T
		\end{bmatrix}^T
		$}
	\nomenclature[33]{\(\mu\)}{Scalar weight used for regularization}
	\nomenclature[34]{\(\hat{H}_l\in\mathbb{R}^{n\times n}\)}{Hierarchical Hessian of level $l$}
	\nomenclature[35]{\(R_l\in\mathbb{R}^{n\times n}\)}{Factor of $\hat{H}_l$ such that $\hat{H}_l = R_l^TR_l$}

	\printnomenclature[3cm]
	\label{sec:nomenclature}

	\subsection{Kinematic control}
	
	In kinematic control we want to find a minimizer to a given non-linear kinematic error ${e}_2({q(t)})\in\mathbb{R}^{m_2}$ while not violating the constraint ${e}_1({q(t)}) \in\mathbb{R}^{m_1}\leqq 0$ 
	\begin{align}
	\mini_{q(t)} \quad & {e}_2({q(t)}) \coloneqq {f}_{2,d}(t) - {f}_2({q(t)})	\label{eq:geomErrFct}\\
		\mbox{s.t.} \quad & {e}_1({q(t)}) \leqq 0 \nonumber
	\end{align}
	The indices $2$ and $1$ symbolize the prioritization between the objectives and the constraints respectively.
	Both equalities and inequalities are encapsulated in the symbol $\leqq$.
	The vector ${q}(t)\in\mathbb{R}^n$ ($n$: number of joints) represents the kinematic configuration of the robot's joints and free-flyer base if any ($\in\text{SE}(3)$, $n_f$: number of free-flyer joints). It is continuous in time $t$.
	$f:\mathbb{R}^n \rightarrow \mathbb{R}^{m}$ is a kinematic function with sufficient continuity properties representing the non-linear kinematics of the robot. $m$ is the task dimension. $f_d(t)\in\mathbb{R}^m$ is the desired value. We define the Jacobian ${J}(q(t)) \coloneqq \nabla_{{q(t)}} {f(q(t))}\in\mathbb{R}^{m\times n}$ so $\nabla_{{q(t)}} {e}(q(t)) = -{J}(q(t))$. 
	
	Non-linear optimization methods with global convergence properties have been proposed in order to find a minimizer to non-linear optimization problems (NLP) like~\eqref{eq:geomErrFct} for some given norm~\cite{Yenamandra2019, Dai2019}.   
	 However, these methods are usually computationally heavy and cannot be applied in real-time control. Instead, one can iteratively (i.e. in every consecutive discrete control step $k$ at time $t_k$) linearize the non-linear kinematic error in time $t$ around the current point $t_k$ with $\{q_k,\dot{q}_k,\ddot{q}_k\}$ and solve an easier since linear problem.
	This can be factually achieved by the second-order time derivative of $e(q(t))$ at time $t_k$ which allows us to define motion controllers linear in the joint accelerations $\ddot{q}_k$
	\begin{equation}
	{\ddot{e}}^{\text{ctrl}}_{\text{PD},k} = -{J_k}{\ddot{q}_{k}} - {\dot{J}_k}{\dot{q}_{k}}
	\label{eq:PDacc}
	\end{equation}
	
	Thereby, ${\ddot{e}}^{\text{ctrl}}_{\text{PD}}$ ($\text{ctrl}$: control) is a PD controller of the form 
	\begin{equation}
	{\ddot{e}}^{\text{ctrl}}_{\text{PD},k} \coloneqq -k_p{e}_k-k_v{\dot{e}_k}
	\label{eq:PDcontroller}
	\end{equation}
    with positive proportional and derivative gains $k_p$ and $k_v$. In the case of velocity based control the first-order derivative suffices with
 	\begin{equation}
	 {\dot{e}}^{\text{ctrl}}_{\text{P},k} = -{J_k}{\dot{q}_{k}}
	 \label{eq:Pvel}
	 \end{equation}
	 and the proportional controller
	 \begin{equation}
	 {\dot{e}}^{\text{ctrl}}_{\text{P},k} \coloneqq -k_p{e}_k
	 \label{eq:Pcontroller}
	 \end{equation}   
    The joint accelerations $\ddot{q}_k$ (or velocities $\dot{q}_k$) resulting from~\eqref{eq:PDacc} or~\eqref{eq:Pvel} (for example as a least-squares solution with fixed $\dot{q}_k$ and ${q}_k$) consecutively let the robot converge to a local solution of~\eqref{eq:geomErrFct}. In sections~\ref{sec:minNonLinGeoFct} and~\ref{sec:fromOptToCtrl} we further detail how this approach can be linked to optimization and control.
    
   	\subsection{Dynamic control}
    
    The motions at a time $t_k$ resulting from the kinematic tasks should obey the instantaneous Newton-Euler equations (or forward dynamics (FD))
    \begin{equation}
    {M}({q_k}){\ddot{q}_k} + {N({q_k},{\dot{q}_k})} = {S}^T{\tau_k} + {J}_c^T(q_k) {\gamma},
    \label{eq:eqOfMotion}
    \end{equation}
    in order to be physically feasible, especially with respect to dynamic limits of the robot.
    This includes limits on the joint torques $\tau\in\mathbb{R}^{n}$ and the contact forces $\gamma\in\mathbb{R}^{n_c}$ ($n_c$: number of contacts). ${S}\in\mathbb{R}^{n_f\times n}$ is a selection matrix to exclude the unactuated free-flyer. ${M}({q})\in\mathbb{R}^{n\times n}$ is the whole-body inertia matrix. ${N({q},{\dot{q}})}\in\mathbb{R}^n$ combines Coriolis, centrifugal, gravitational and frictional force effects. ${J}_c\in\mathbb{R}^{n_c\times n}$ is the contact points' Jacobian matrix.
    
    Note that the original non-linear dynamics are thereby linearized by instantiating the equation at a time $t_k$ and keeping $q_k$ and $\dot{q}_k$ fixed in each respective iteration.
	
	\subsection{Hierarchical control}

	We can combine both the linearized kinematic tasks and the equation of  motion in a hierarchical least-squares problem with $p$ levels as done in~\cite{DeLasa2010,Saab2013,Escande2014,Herzog2016}. 
	This boils down to solving a sequence of linear least-squares programs (in the sense of sequential hierarchical least-squares programming (S-HLSP) and assuming that we want to find a local minimizer of the task error~\eqref{eq:geomErrFct} in the 2-norm) for $l=1\cdots p$
	\begin{align}
	\min_{{x_k},{w}_{k,l}} \quad & \frac{1}{2} \|{w}_{k,l}\|^2\qquad\qquad l=1 \cdots p     \label{eq:generalHierarchy}\\
	\mbox{s.t.} \quad & {A}_{k,l}{x} + {b}_{k,l} \hspace{3pt} \leqq \hspace{3pt} {w}_{k,l} \nonumber\\
	& {\underline{A}}_{k,l-1}{x} + {\underline{b}}_{k,l-1} \hspace{3pt} \leqq \hspace{3pt} {\uw}_{k,l-1}^{*} \nonumber
	\end{align}
	This results in a new instantaneous robot state 
	\begin{equation}
	{x}_k =
	\begin{bmatrix}
	{\ddot{q}}_k^{T} & {\tau}_k^{T} & {\gamma}_k^{T}
	\end{bmatrix}^T,
	\end{equation} 
	at a given discrete control time step $t_k$. 
	${A}$ and ${b}$ are determined by the linearizations of the equation of motion~\eqref{eq:eqOfMotion} or the kinematic tasks~\eqref{eq:PDacc}. Priorities can be conveniently chosen such that real-world constraints are respected without compromises as is the case for weight-based constrained optimization ($p=2$)\cite{vaillant:auro:2016,pfeiffer2017}. ${w}_i\in\mathbb{R}^{m}$ is a slack variable which relaxes infeasible objectives for example due to task conflict.
	${\uw}_{l-1}^{*}$ are the optimal values obtained from solving the problems for $i<l$. 
	
	\subsection{Kinematic and algorithmic singularities}
	
	As it is, the above hierarchy~\eqref{eq:generalHierarchy} cannot intrinsically deal with kinematic singularities of the Jacobians $A\coloneqq J$ in~\eqref{eq:PDacc}. More, objectives on different priority levels might conflict with each other resulting in algorithmic singularities. Therefore, appropriate singularity resolution methods have to be considered in order to prevent unstable robot behavior due to numerics.

	\section{The hierarchical Newton's method}
	\label{sec:minNonLinGeoFct}
	
	As has been proposed by~\cite{Deo:1993}, Newton's method can be used to solve non-linear inverse kinematics problems~\eqref{eq:geomErrFct} in a least-squares sense. This maintains critical second-order information in the vicinity of singularities of the Jacobian of the kinematic error~\eqref{eq:geomErrFct} (see bellow). A similar reasoning is applicable for prioritized inverse kinematic control as observed in~\cite{Pfeiffer2018}, where only a hierarchical Quasi-Newton method was proposed. Now, we derive the hierarchical Newton's method. We start with the following non-linear hierarchical least-squares problem (NL-HLSP, the time dependency $t$ is omitted to ease readability)
	\begin{align}
	\min_{{q},{w}_l} \quad & \frac{1}{2} \left\|{w}_l\right\|^2 \qquad\quad\hspace{8pt} l=1\cdots p
	\label{eq:NL-HLSP}\\
	\mbox{s.t.} \quad & {e}_l({q}) \hspace{3pt} \leqq \hspace{3pt} {w}_l \nonumber\\
	& {\ue}_{l-1}({q}) \hspace{3pt} \leqq \hspace{3pt} {\uw}_{l-1}^* \nonumber
	\end{align}
	The goal is to minimize the constraint violation ${w}_l$ of each level $l$ `at best' (in a least-squares sense). Already obtained optimal violations of previous levels ${\uw}_{l-1}^*$ must stay unchanged. 
	
	In the sense of the active-set method, we formulate the Lagrangian of an equality only problem of~\eqref{eq:NL-HLSP} at level $l$ 
	\begin{align}
	\mathcal{L}_l = \frac{1}{2}{w}_l^T{w}_l + {\lambda}_{l,l}^T({w}_l - {e}_l) + {\ulambda}_{l-1,l}^T({\uw}_{l-1}^* - {\ue}_{l-1})
	\label{eq:lagrangianGenHier}
	\end{align}
	Only active constraints with ${e}_l \geq 0$ and ${\ue}_{l-1} = {\uw}_{l-1}^*$ are considered.
	${\lambda}_{l,l}$ and ${\ulambda}_{l-1,l}$ are the Lagrange multipliers associated with the active constraints of level $l$ and the previous levels $1,\dots,l-1$, respectively.
	%Note that ${\lambda}\in\mathbb{R}^{\sum_{i=1}^p m_i,p}$ is a matrix with $p$ columns.
	
	The non-linear first order optimality conditions are
	\begin{align}
	\nabla_{{q},{w}_l,{\lambda}_{l,l},\ulambda_{l-1,l}} \mathcal{L}_l &= {K}_l({q},{w}_l,{\lambda}_{l,l}, {\ulambda}_{l-1,l}) 	\label{eq:optcondconstrained}\\
	&= \protect\begin{bmatrix} {J}_l^T {\lambda}_{l,l} + {\uJ}_{l-1}^T {\ulambda}_{l-1,l} \\ {w}_l + {\lambda}_{l,l} \\ {w}_l - {e}_l \\ {\uw}_{l-1}^* - {\ue}_{l-1} \protect\end{bmatrix} = {0}
	\nonumber
	\end{align}
	We perturb the KKT system (not in time $t$ but only in configuration space $q$, see Sec.~\ref{sec:fromOptToCtrl}) leading to the Newton step 
	\begin{equation}
	{K}_l({x_k} + {\Delta x_k}) = {K}_l({x_k}) + \nabla{K}_l({x_k}) {\Delta x_k} = {0}
	\label{eq:newtonstep}
	\end{equation} 
	Here, the index $k$ indicates the current iteration of the Newton's method.
	The variable vector $x$ (and its increment $\Delta x$) is given by
	\begin{equation}
	{x}^T = \begin{bmatrix}
	{q}^T & {w}_l^T & {\lambda}_{l,l}^T & {\ulambda}_{l-1,l}^T
	\end{bmatrix}
	\end{equation}
	The Lagrangian Hessian is
	\begin{align}
	\nabla{K}_l({x}) = \protect\begin{bmatrix} {\hat{H}}_l & {0} & {J}_l^T & \underline{J}_{l-1}^T \\ 
	{0} & {I} & {I} & 0\\ 
	{J}_l & {I} & {0} & 0 \\
	\underline{J}_{l-1} & 0 & 0 & 0
	\protect\end{bmatrix}
	\end{align}
	We refer to the expression
	\begin{equation}
	\label{eq:hessianHier}
	\nabla^2_q\mathcal{L}_l = {\hat{H}}_l = \sum_{d=1}^{m_l} \lambda_{l,l,d} {H}_{l,d} + \sum_{i=1}^{l-1}\sum_{d=1}^{m_i} \lambda_{i,l,d} {H}_{i,d}
	\end{equation}
	as the hierarchical Hessian. $H_l\coloneqq\nabla_q^2 f_l$ are the Hessians of the respective kinematics $f_l(q)$.
	
	Equation~\eqref{eq:newtonstep} is also the optimality condition of the HLSP
	\begin{align}
	\label{eq:optCondQP}
	\min_{\Delta q_k, w_{k,l}} & \frac{1}{2} \left\|w_{k,l}\right\|^2 + \frac{1}{2} \Delta q_k^T {\hat{H}}_{k,l} \Delta q_k \\
	s.t. \quad & e_{k,l} + J_{k,l} \Delta q_k = w_{k,l} \nonumber\\
	& \underline{e}_{k,l-1} + \uJ_{k,l-1} \Delta q_k = \uw_{k,l-1}^*\nonumber
	\end{align}
	which we refer to as the hierarchical Newton's method in combination with the active-set method. As we repeatedly execute the above steps at the current iterate $q_k$, a non-linear hierarchical least-squares programming is turned into a linear one and solved until convergence; we can also refer to this method as sequential hierarchical least-squares programming (S-HLSP).
	If ${\hat{H}}_l$ is positive definite, 
	its Cholesky decomposition ${\hat{H}}_l = {R}_l^T{R}_l$ exists and we get the least-squares program
	\begin{align}
	\label{eq:Nopt}
	\min_{{\Delta q_k}} \quad
	&\frac{1}{2} \left \Vert
	\begin{bmatrix}
	{J}_{k,l}\\
	{R}_{k,l}
	\end{bmatrix}
	{\Delta{q}_k} -
	\begin{bmatrix}
	{e}_{k,l}\\
	{0}
	\end{bmatrix}
	\right \Vert^2_2 \qquad l=1\cdots p
	\\
	\text{s.t.} \quad &
	{\ue}_{k,l-1} - {\uJ}_{k,l-1}  {\Delta q_k}  = {\uw}_{k,l-1}^{*}\nonumber
	\end{align}
	If ${\hat{H}}_l$ is neglected, we obtain the GN algorithm. If the Jacobian $J_l$ is rank deficient, the problem is ill-posed and results in a numerically unstable solution $\Delta q_k$.
	If $\hat{H}_l = \mu^2 I$ is chosen as an identity matrix with weight $\mu$, we get the LM algorithm. If $\hat{H}_l$ is approximated by the BFGS algorithm~\cite{Pfeiffer2018} using the gradient $\nabla_q \mathcal{L}$ in~\eqref{eq:optcondconstrained}, we get a Quasi-Newton method.
	
	The hierarchical Newton's method is in the form of the hierarchical least-squares program formulated in~\eqref{eq:generalHierarchy} (with $w_l$ given implicitly).
	Efficient solvers to~\eqref{eq:Nopt} based on the active-set method are described in e.g.~\cite{Escande2014} or~\cite{dimitrov:2015}. 
	
%	\subsection{Incorporation of inequality constraints}
%	\label{eq:hierHineq}
	Inequality constraints are incorporated by means of active and inactive constraints and due to the problem formulation with slack variables, see~\cite{Kanoun2011}. Above first order optimality conditions extend to the Karush-Kuhn-Tucker conditions. Inactive constraints $i$ thereby result in ${w}_i = {0}$, ${\lambda}_{j,i} = {0}$, not further influencing the Hessian ${\hat{H}}_j$ on some level $j \geq i$.
	
	%\subsection{Computation of the hierarchical Hessian and handling indefiniteness}
	
	For the hierarchical Hessian calculation ${\hat{H}}_l$ of level $l$~\eqref{eq:hessianHier} we need to calculate the second order derivatives
	\begin{equation}
	{H} = \nabla^2_{{q}} {f}({q})
	\label{eq:analyticHessian}
	\end{equation}
	of the kinematic functions ${f}({q})$ for all levels $1$ to $l$.
	For this we follow~\cite{Erleben2017} with computational complexity of $\mathcal{O}(n^2)$.
	
	Since the hierarchical Hessian ${\hat{H}}$ can become indefinite, the Cholesky decomposition for obtaining the factor $R$ can not be applied. We use the symmetric Schur decomposition ($O(9n^3)$~\cite{Golub1996}) to obtain the spectral decomposition ${\hat{H}}=QUQ^{T}$ with $U = \diag(\lambda({\hat{H}}))$. Negative eigenvalues in $U$ are then replaced by a small positive threshold such that $R=\sqrt{U}Q^T$ and convexity of the optimization problem is maintained. This method proves to be faster than other regularization methods like~\cite{Higham1986} which is based on the (asymmetric) SVD decomposition ($O(12n^3)$~\cite{Golub1996} plus an additional Cholesky decomposition $O(n^3/3)$ in order to obtain $R$). A positive definite approximation of the Hessian can also be obtained by the BFGS algorithm as detailed in~\cite{Pfeiffer2018}.

	\section{From optimization to kinematic control}
	\label{sec:fromOptToCtrl}

	In Section~\ref{sec:minNonLinGeoFct} we introduced the Newton's method of constrained optimization to bring a non-linear kinematic error to zero. The Newton's method corresponds to a quadratic Taylor approximation~\cite{Deo:1993}
	around ${q}$ that is valid within a neighborhood (called trust region) of it, and allows for a bounded step ${\Delta q}$. Assuming a zero order holder with a control time step of $\Delta t=1$~s; the change of joint configuration is ${\Delta q} = \Delta t{\dot{q}} = {\dot{q}}$. This allows us to make a trivial connection between optimization, aiming to make a step ${\Delta q}$ towards the optimum (note how the KKT system is not disturbed in time $t$ in the derivation of the hierarchical Newton's method~\eqref{eq:newtonstep}), and control, aiming to determine the next robot state triple $\{{q}_k,{\dot{q}_k},{\ddot{q}_k}\}$. It is also noteworthy that in this case the linearization in time $t$ as done in~\eqref{eq:PDacc} corresponds to the Gauss-Newton algorithm and its subsequent numerical instability at kinematic singularities.
	
	However, usually robots are controlled at much higher rates, that is $\Delta t \ll 1$~s. Yet, the time step $\Delta t$ connects the two entities of `optimization' (optim) and `control' (assuming a simple proportional controller $\dot{e}^{\text{ctrl}}_{\text{P},k}$ for now)
	\begin{align}
	{w}^{\text{optim}}_k \coloneqq {J}_k{\Delta q_k} + \Delta t {\dot{e}}^{\text{ctrl}}_{\text{P},k} = \Delta t({J}_k {\dot{q}}_k + {\dot{e}}_{\text{P},k}^{\text{ctrl}}) 
	\eqqcolon \Delta t {w}^{\text{ctrl}}_k
	\label{eq:wscale}
	\end{align}
	That is, we calculate a new velocity ${\dot{q}}_k$ but only make a step ${\Delta q}_k = \Delta t {\dot{q}}_k$ towards the optimum.
	Consequently, the model needs to be updated with the Hessian of the Lagrangian \eqref{eq:lagrangianGenHier} using  ${w}^{\text{optim}}_k$ and ${\lambda}^{\text{optim}}_k$. Since solving the constrained control problems~\eqref{eq:Ncontrol} yields ${w}^{\text{ctrl}}_k$ and ${\lambda}^{\text{ctrl}}_k$, a scaling of the form ${w}^{\text{optim}}_k = \Delta t {w}^{\text{ctrl}}_k$ according to \eqref{eq:wscale} is required. Due to the linear dependency between the slack ${w}$ and the Lagrange multipliers ${\lambda}$~\cite{dimitrov:2015} we further get ${\lambda}^{\text{optim}}_k = \Delta t {\lambda}^{\text{ctrl}}_k$.
	In what follows, we write ${w} = {w}^{\text{optim}}$ and ${\lambda} = {\lambda}^{\text{optim}}_k$.
	
	With these considerations in mind we can directly derive the hierarchical Newton's method of control from~\eqref{eq:Nopt} (with $k$ now again representing the time $t_k$ and not the $k$-th Newton iteration)
	\begin{align}
	\label{eq:Ncontrol}
	\min_{{\dot{q}_k}} \quad&
	\frac{1}{2} \left \Vert
	\begin{bmatrix}
	{J}_{k,l}\\
	{R}_{k,l}
	\end{bmatrix}
	{\dot{q}_{k}} +
	\begin{bmatrix}
	{\dot{e}}^{\text{ctrl}}_{\text{P},k,l}\\
	{0}
	\end{bmatrix}
	\right \Vert^2_2\qquad\text{for }l=1\cdots p
	\\
	\text{s.t.} \quad & -{\dot{\ue}}_{k,l-1}^{\text{ctrl}} - {\uJ}_{k,l-1} {\dot{q}_k}   \leqq   {\uw}_{k,l-1}^{*}\nonumber
	\end{align}
	Neglecting the second order information $\hat{H}_l$ (or its Cholesky factor $R_l$ thereof) yields the hierarchical GN algorithm of control.
	If the Jacobian $J_l$ is rank deficient the problem is ill-posed and results in a numerically unstable solution $\dot{q}_k$.
	If $\hat{H}_l = \mu^2 I$, $I$ being the identity matrix with weight $\mu$, we get the LM algorithm. If $\hat{H}_l$ is approximated by the BFGS algorithm~\cite{Pfeiffer2018}, by using the gradient $\nabla_q \mathcal{L}$ in~\eqref{eq:optcondconstrained}, the slacks $w^{\text{ctrl}}$ and the Lagrange multipliers $\lambda^{\text{ctrl}}$, we get the Quasi-Newton method of control.
%	Inequality constraints can be handled as described in sec.~\ref{eq:hierHineq}.
	The trust region constraint is converted into a limit on the joint velocities.
	In the following we refer to Newton's method as being the `augmented' (as in augmented with second order information) version of the GN algorithm~\cite{Dennis1981}.
	
	As described in~\cite{Pfeiffer2018} we switch between the GN algorithm and the Newton's method judging upon the residual of the GN algorithm
	\begin{equation}
    \frac{1}{2}\Vert {J}_{k,l} {\dot{q}_k} + {\dot{e}}_{\text{P},k,l}^{\text{ctrl}}\Vert^2_2 < \nu
	\end{equation}
	By doing so, the joints are `freed' from the full rank second order augmentation whenever the linearization represents the non-linear original task function sufficiently enough. This way we ensure the best possible convergence of lower priority levels. Note that the threshold $\nu = 10^{-12}\Delta t^2$ is now dependent of the time step $\Delta t$ in accordance with~\eqref{eq:wscale}.

	\section{Dynamically feasible kinematic control}
	\label{sec:fromInvKintoInvDyn}
	In previous Sec.~\ref{sec:fromOptToCtrl} we formulated the GN algorithm and Newton's method of control only in the velocity domain. In this section, we extend our approach to equation of motion~\eqref{eq:eqOfMotion} of second order (dynamics). In Sec.~\ref{sec:dampAcc} we argue why Newton's method cannot be straightforwardly extended to the acceleration domain. Note that this also concerns regularization of acceleration based tasks as is commonly done in robotics (albeit with a small weight), see for example~\cite{vaillant:auro:2016, Herzog2016}. Therefore, our idea for acceleration-based control is to change the controller ${\dot{e}}^{\text{ctrl}}_{\text{P}}$ from a linear proportional controller to some controller ${\dot{e}}^{\text{ctrl}}_{\text{PD}}$ that emulates PD control ${\ddot{e}}^{\text{ctrl}}_{\text{PD}}$ in the velocity domain. In Sec.~\ref{sec:fromVelToAcc} we show how this can be achieved and applied to the equation of motion (see Sec.~\ref{sec:inclDyn}).
	
	\subsection{Damping in acceleration-based control}
	\label{sec:dampAcc}
	In the following we show that the instantaneous acceleration based kinematic optimization at a discrete time $t_k$ 
	\begin{align}
	&\min_{{\ddot{q}_k}} \quad
	\frac{1}{2} \left \Vert
	\begin{bmatrix}
	{J_k}\\
	{R_k}
	\end{bmatrix}
	{\ddot{q}_k} +
	\begin{bmatrix}
	{\dot{J}_k}{\dot{q}_k} + {\ddot{e}}^{\text{ctrl}}_{\text{PD},k}\\
	{0}
	\end{bmatrix}
	\right \Vert^2_2
	\label{eq:dampedAccCtrl}
	\end{align}
	leads to low frequency oscillations over time $t$. 	
	Solving~\eqref{eq:dampedAccCtrl} using the pseudo-inverse to obtain the instantaneous joint accelerations and using Euler integration to obtain the new joint velocities at time $t_{k+1} = t_k + \Delta t$ we get (assuming full rank of $\protect\begin{bmatrix} J_k^T & R_k^T \protect\end{bmatrix}^T$)
	\begin{align}
	\dot{q}_{k+1} &= (J_k^TJ_k + R_k^TR_k)^{-1}\\
	&((R_k^TR_k + (1 - \Delta t k_v)J_k^TJ_k)\dot{q}_k - \Delta t J_k^T(k_pe_k + \dot{J}_k\dot{q}_k))\nonumber\\
	q_{k+1} &= q_k + \Delta t \dot{q}_k	
	\label{eq:eulerAccDamp}
	\end{align}
	It can be observed that the positive definite augmentation $R^TR$ reduces the negative definite term $-\Delta t k_vJ^TJ$ (which represents the derivative term of the PD controller) and therefore leads to an under-damped system with characteristic overshooting behaviour. Damping in the form of ${R} = \mu {I}$ is commonly applied in robotics for regularization purposes~\cite{vaillant:auro:2016, Herzog2016}. The weight $\mu$ is usually chosen small such that the oscillations are small in amplitude and do not negatively influence practical task achievement. From a theoretical point of view this still might be undesirable.
	
	In order to illustrate the above derivations we assume a point mass with a continuous 1D translational degree of freedom (DoF) $q(t)$ and $f(t)=q(t)$. The desired position is $q_d = 0$. The task error is then $e(t) = q_d - q(t) = -q(t)$. The Jacobian of this robot is $J = \frac{df}{dq} = \frac{dq}{dq} = 1$, the time derivative of the Jacobian is $\dot{J} = 0$.
	
	The acceleration based controller~\eqref{eq:PDacc} becomes	
	\begin{align}
	\ddot{q}(t) + k_v\dot{q}(t) + k_p q(t) &= 0 \label{eq:1dacc}
	\end{align}
	with the solution
	\begin{equation}
	q(t) = \exp{-\delta t} (C \cos(w_d t) + D \sin(w_d t))
	\label{eq:2ndOrderOdeSolution}
	\end{equation}
	where $\delta=k_v/2/m$ and $w_d = \sqrt{\delta^2 - k_p/m}$. $C$ and $D$ are constants of integration.
	Critical damping with exponential convergence can be achieved for $k_v(k_p) =  2\sqrt{mk_p}$ such that $w_d=0$.
	
	\begin{figure}[t!]
		\centering{\includegraphics[width=1\columnwidth]{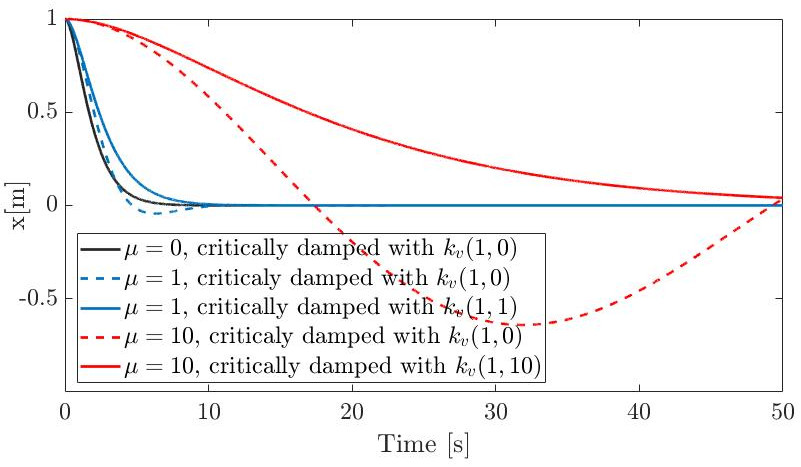}}
		\caption{Equation~\eqref{eq:2ndOrderOdeSolution} plotted for $m=1$, $k_p=1$, different $\mu$ and $k_v(k_p) =  2\sqrt{mk_p}$ (black line and dashed lines) or $k_v(k_p,\mu) = 2\sqrt{m(1+\mu^2)k_p}$.}
		\label{fig:accBasedCriticalDampingCurves}
	\end{figure}
	
We now apply a damping term $\mu$ to the system~\eqref{eq:1dacc}
	\begin{align}
	\begin{bmatrix}
	1\\
	\mu
	\end{bmatrix}
	\ddot{q}(t)
	+
	\begin{bmatrix}
	k_v\dot{q}(t) + k_pq(t)\\
	0
	\end{bmatrix}
	=
	\begin{bmatrix}
	0\\0
	\end{bmatrix}
	\end{align}
	We can convert this system into a homogenous ODE with the same solution as in~\eqref{eq:2ndOrderOdeSolution} by applying the pseudo-inverse
	\begin{equation}
	\ddot{q}(t) + \frac{k_v}{1+\mu^2}\dot{q}(t) + \frac{k_p}{1+\mu^2}q(t) = 0
	\end{equation}
	The influence of the damping $\mu$ on the critically damped task gains $k_p$ and $k_v$ is clearly exposed. Critical damping can be achieved with 
	\begin{equation}
	k_v(k_p,\mu) = 2\sqrt{m(1+\mu^2)k_p}
	\end{equation}
	Some convergence curves for $q(0) = 1$, $\dot{q}(0) = 0$, $\Delta t = 5$~ms are plotted in Fig.~\ref{fig:accBasedCriticalDampingCurves}. Note that the damping $\mu$ influences the critically damped system negatively (i.e. overshooting) if the gain is chosen according to $k_v(k_p)$ instead of $k_v(k_p,\mu)$.
	
	For a complicated 3D robot with more and especially coupled DoF's, and a varying ${R}(q,\mu)$, it seems cumbersome to find the expression for critical damping $k_v(k_p,{R})$ such that overshooting behavior can be prevented. 
	Therefore, we favour to shift the whole problem into the velocity domain and emulate acceleration-based control by formulating an appropriate controller.

	\subsection{Acceleration control expressed in the velocity domain}
	\label{sec:fromVelToAcc}
	
	We propose a controller ${\dot{e}}^{\text{ctrl}}_{\text{PD}}$ that is able to emulate acceleration based PD control in the velocity domain
		\begin{equation}
	{\dot{e}}^{\text{ctrl}}_{\text{PD},k} \coloneqq -\dot{e}_k - 
	\Delta t({\ddot{e}}^{\text{ctrl}}_{\text{PD},k}+{\dot{J}_k}{\dot{q}_k})
	\label{eq:PDvel}
	\end{equation}
	If we use this controller in the unconstrained Newton's method of control~\eqref{eq:Ncontrol}, by application of the pseudo-inverse we get 
	(note that $\dot{e}^{\text{ctrl}}_{\text{PD},k}$ 
	already incorporates the Euler integration $\dot{q}_{k+1} = \dot{q}_k + \Delta t \ddot{q}_k$; 
	the new decision variable in~\eqref{eq:Ncontrol} is therefore $\dot{q}_{k+1}$ instead of $\dot{q}_{k}$)
	\begin{align}
	\dot{q}_{k+1} &= (J_k^TJ_k + R_k^TR_k)^{-1}\label{eq:eulerabio1}\\
	&((1 - \Delta tk_v)J_k^TJ_k\dot{q}_k - \Delta tJ_k^T(k_pe_k + \dot{J}_k\dot{q}_k))\nonumber\\
	q_{k+1} &= q_k + \Delta t \dot{q}_k	
	\label{eq:unconN}
	\end{align}
	In contrast to~\eqref{eq:eulerAccDamp} we can now see that the derivative term $-\Delta tk_vJ_k^TJ_k$ is not influenced by the augmentation $R_k^TR_k$. As long as the system is critically damped $k_v(k_p) = 2\sqrt{mk_p}$ and $k_v > 1/\Delta t$ exponential convergence is ensured. If there is no augmentation ($R_k=0$, GN algorithm) we get the original acceleration based PD control dynamics
	\begin{align}
	\dot{q}_{k+1} &= \dot{q}_k + \Delta tJ_k^+(- k_vJ_k\dot{q}_k - k_pe_k - \dot{J}_k\dot{q}_k)\label{eq:eulerpd}\\
	q_{k+1} &= q_k + \Delta t \dot{q}_k	
	\label{eq:1ddotepdctrl}
	\end{align}
	It can be seen that the augmented system~\eqref{eq:eulerabio1} exhibits slower convergence than the equivalent undamped acceleration based PD controller~\eqref{eq:eulerpd} depending on the magnitude of $R_k^TR_k\geq 0$ since
	$\Vert(J_k^TJ_k + R_k^TR_k)^{-1}J_k^TJ_k\dot{q}_k\Vert \leq \Vert \dot{q}_k \Vert$. However, we can argue that we only use Newton's method when slow joint movements in singular configurations are desirable anyways and the particular type of critically damped PD motions is not further relevant.
	
	\begin{figure}[t!]
		\centering{\includegraphics[width=1\columnwidth]{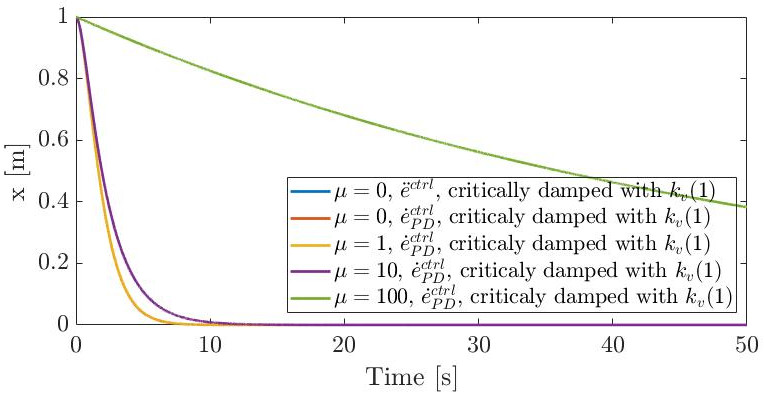}}
		\caption{Plots of convergence of the 1-D mass governed by~\eqref{eq:1ddotepdctrl}. The larger $\mu$ the slower the robot behaves. Simultaneously, the difference to the control of the undamped ($\mu=0$) acceleration based PD controller $\ddot{e}^{\text{ctrl}}$ increases. However, exponential convergence is always achieved.}
		\label{fig:dote_pd_ctrl}
	\end{figure}
	
	To come back to the 1-D robot example from the previous section~\ref{sec:dampAcc} we get
	\begin{align}
	\dot{q}_{k+1} &=  \frac{1}{1 + \mu^2}
	(1 - \Delta tk_v\dot{q}_k - \Delta tk_pe_k)\nonumber\\
	q_{k+1} &= q_k + \Delta t \dot{q}_k	
	\end{align}
	Some convergence curves for $q(0)=1$, $\dot{q}(0) = 0$, $\Delta t = 5$~ms are plotted in Fig.~\ref{fig:dote_pd_ctrl}. With increasing damping $\mu$ the robot convergence becomes slower and diverges from the undamped control generated by the acceleration based PD controller $\ddot{e}^{\text{ctrl}}$. Unlike the acceleration based controller $\ddot{e}^{\text{ctrl}}$ however, exponential convergence is achieved by $\dot{e}^{\text{ctrl}}_{\text{PD}}$ for any $\mu$ as long as the system is critically damped with $k_v = 2\sqrt{mk_p}$.
	
 At the optimum of each instantaneous control cycle we solve an equality only least-squares problem projected into the null-space basis of the active constraints of the higher priority levels~\cite{dimitrov:2015}. Therefore the same argumentation as above using the pseudo-inverse~\eqref{eq:unconN} holds for the constrained Newton's method of instantaneous control.
	
The above scheme uses linear integration. This forces to adopt Euler angles to represent 3D rotations such as the robot base orientation. To avoid gimbal lock, we use them as a local parametrization: a rotation $Q$ is written $Q_0 Q(\theta)$ with $Q_0$ fixed and $Q(\theta)$ the Euler parametrization. After each control iteration $Q_0$ is set to $Q_0 Q(\theta)$ and $\theta$ is set to $0$. We assume small changes of $\theta$ between iterations.
	
	\subsection{Including the dynamics}
	\label{sec:inclDyn}
	We have formulated our second order motion controllers in the velocity domain. Similarly, the acceleration components of the equation of motion~\eqref{eq:eqOfMotion} are replaced by finite differences
	\begin{equation}
		\ddot{q}_k = \frac{\dot{q}_{k+1} - \dot{q}_{k}}{\Delta t}
		\label{eq:AccToVelForwardDifference}
	\end{equation}
    such that we get
	\begin{align}
	&\begin{bmatrix}
	{M}(q_k) & -{S}^T & -{J}^T_{c}(q_k)
	\end{bmatrix}
	\begin{bmatrix}
	{\dot{q}}_{k+1}\\
	\Delta t{\tau}_k\\
	\Delta t{\gamma}_k
	\end{bmatrix}
	= {M}{\dot{q}_k} - \Delta t{N({q_k},{\dot{q}_k})}
	\label{eq:intEqOfMot}
	\end{align}
	For numerical robustness it is desirable to keep the conditioning of the system matrix $\protect\begin{bmatrix} {M}(q_k) & -{S}^T & -{J}^T_{c}(q_k)\protect\end{bmatrix}$ so we compute ${\dot{q}}_{k+1}$, $\Delta t{\tau}_k$ and $\Delta t{\gamma}_k$.
	
	The equation of motion can be considered full rank if the inertia matrix ${M}$ is physically consistent and therefore positive definite~\cite{Udwadia1992,Udwadia2010}. This means that the system matrix of the equation of motion
	$
	\begin{bmatrix}
	{M}(q_k) & -{S}^T & -{J}^T_{c}(q_k)
	\end{bmatrix}
	$
	is not concerned with kinematic singularities of the contact Jacobians ${J}_c(q_k)$ due to the full-rank property of $M(q_k)$ (`row rank equals column rank'). Furthermore, the equation of motion is already linear in the accelerations $\ddot{q}_k$ (or velocities $\dot{q}_{k+1}$ in case of the forward integration and only non-linear in $\dot{q}_k$), joint torques $\tau_k$ and generalized contact forces $\gamma_k$. Hence, a Taylor expansion for the purpose of linearization is not necessary. We therefore do not consider the equation of motion in the Hessian calculation of the lower level linearized constraints.
	
	\section{Validation}
	\label{sec:validation}

	Here, we assess our proposed method named LexDynAH (acronym for \textbf{Lex}icographic Augmentation for \textbf{Dyn}amics with \textbf{A}nalytic \textbf{H}essian).
	We aim to validate that we can achieve: (i) agreement between the velocity and acceleration based controllers ${\dot{e}}^{\text{ctrl}}_{\text{PD}}$~\eqref{eq:PDvel} 
	and ${\ddot{e}}^{\text{ctrl}}_{\text{PD}}$~\eqref{eq:PDcontroller}; and (ii) numerical stability in singular robot configurations.
	First, we conduct three simulations with simple 2D stick robots (sec.~\ref{sec:dynTests}) to confirm our derivations.
	Then, we apply our method in two real HRP-2Kai robot experiments (sec.~\ref{sec:realTests}) that is position controlled. HRP-2Kai has $32$~actuated DoF and a $6$~DoF un-actuated free-flyer. The control frequency is $200$~Hz ($\Delta t = 5$~ms).
	Equation~\eqref{eq:Ncontrol} is solved with the hierarchical least-squares warm-started active set solver LexLSI~\cite{dimitrov:2015}. The trust region limit (Sec.~\ref{sec:fromOptToCtrl}) is chosen as $0.1$~m or~rad with an adaptive heuristic for real-time hierarchical control as in~\cite{Pfeiffer2018}. 
	
	For comparison, we use the hierarchical Quasi-Newton method presented in~\cite{Pfeiffer2018} (referred to as LexDynBFGS). It was originally designed for optimization problems of the form~\eqref{eq:Nopt} but including the modifications presented in this paper it is identical with LexDynAH except that the hierarchical Hessian~\eqref{eq:hessianHier} is approximated by the BFGS algorithm.
	Furthermore, for regularized acceleration based control~\eqref{eq:PDacc} we use the QP solver
	LSSOL~\cite{gill:techrep:1986} ($p=2$). Inequality constraints are only allowed on level~$1$. Since the notion of constraint relaxation is not introduced in LSSOL, the feasibility of these constraints has to be guaranteed in order to avoid solver failures. On both levels a soft hierarchy can be established by weighting tasks against each other (therefore we call this solver {W}eighted {L}east {S}quares - WLS). Such a control setup is commonly found in the literature~\cite{abe:sca:2007,collette:humanoids:2007,feng:humanoids:2013,kuindersma:icra:2014,vaillant:auro:2016,pfeiffer2017}.

	\subsection{Three simulation toy examples}
	\label{sec:dynTests}
	
	\subsubsection{Example~1 -- Freely swinging pendulum}
	\label{sec:freelySwingingPendulum}
	\begin{figure}[ht!]
		\centering
		\begin{tikzpicture}[scale=0.5]
		
		\node[circle,draw=black,inner sep=0pt, outer sep=0pt,thick,fill=white] (joint0) at ($(0,0)$) [] {$q_1$};
		\node[circle,draw=black,inner sep=0pt, outer sep=0pt,thick,fill=white] (joint1) at ($(2,0)$) [] {$q_2$};
		\draw[line width=0.1cm] ($(joint0.east) + (0,0)$) -- ($(joint1.west)$);
		
		\draw[line width=0.1cm] ($(joint1.east)$) -- ($(4,0)$);
		
		\draw[->,line width = 0.05cm] ($(2,1.5)$) -- ($(2,0.5)$) node [midway,right] {${g}$};
		
		\begin{pgfonlayer}{bg} 
		\draw[->,line width = 0.05cm] ($(joint0)$) -- ($(joint0) + (0,1)$) node [midway,above=0.25] {${z}$};
		\draw[->,line width = 0.05cm] ($(joint0)$) -- ($(joint0) + (-1,0)$) node [near end=10,above] {${x}$};
		\end{pgfonlayer} 
		
		\end{tikzpicture}
		\caption[Example~1: initial configuration of the robot.]{\textbf{Example~1}: initial configuration of the robot.}
		\label{fig:2linkrobot}
	\end{figure}
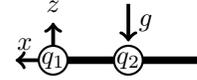
	\begin{figure}[t!]
		\begin{tcolorbox}[title=Hierarchy for example 1,boxsep=1pt,left=2pt,right=3pt,top=2pt,bottom=1pt]
			\begin{enumerate}
				\setcounter{enumi}{0}
				\itemsep0em
				\item[L.$1$] 
				\begin{itemize}
					\itemsep0em
					\item Equation of motion, acceleration-based~\eqref{eq:eqOfMotion} or velocity-based~\eqref{eq:intEqOfMot}
					\item ${\tau} = {0}$
				\end{itemize}
				\item[L.$2$]
				\begin{itemize}
					\item Regularization term ${\dot{q}} = {0}$ (vel. based), expressed in accelerations by using finite differences~\eqref{eq:AccToVelForwardDifference} in case of acceleration-based control
				\end{itemize}
			\end{enumerate}
		\end{tcolorbox}
		\caption[{Example~1}. Hierarchy tasks for swinging pendulum.]{\textbf{Example~1}. Hierarchy for swinging pendulum.}
		\label{fig:example1hierarchy}
	\end{figure}
	
	\begin{figure}[t!]
		\centering
		\centering{\includegraphics[width=1\columnwidth]{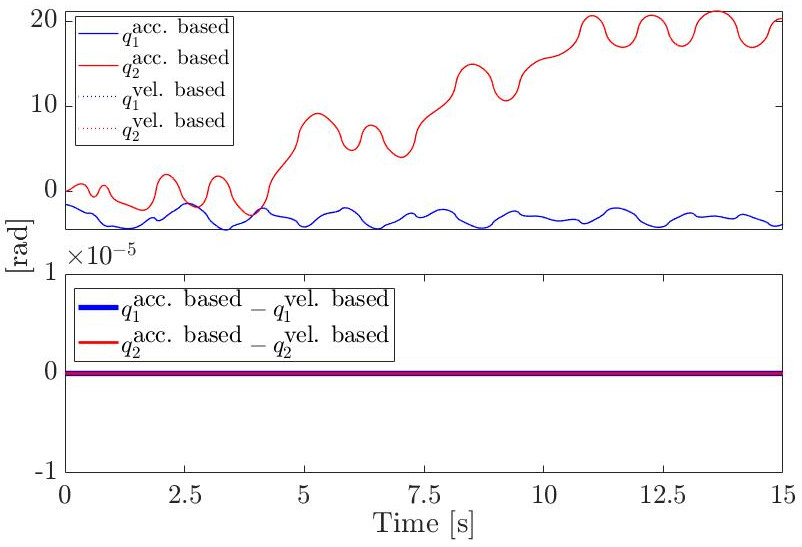}}
		\caption{\textbf{Example~1}. The upper graph shows the joint positions of the swinging pendulum  for the acceleration-based (`acc. based') equation of motion and the velocity-based (`vel. based') ones (full superposition). The lower graph shows the difference between the values.}
		\label{fig:dynTest_pos_equalizedSpecialInt}
	\end{figure}
	
	consists of a fixed-base planar robot with two links (both having unit length and unit mass) and two revolute joints. The joint torques of both joints are set to zero so that it swings freely. The initial configuration is set to $[-\pi/2,0]$~rad with zero velocity (see Fig.~\ref{fig:2linkrobot}). The motion of the pendulum is determined by solving the instantaneous equation of motion~\eqref{eq:eqOfMotion} in a least squares sense. The hierarchical problem is described in Fig.~\ref{fig:example1hierarchy}.
	
	Results of Example~1 show that we can reproduce the behavior of the acceleration equation of motion~\eqref{eq:eqOfMotion} by the use of the velocity domain based one~\eqref{eq:intEqOfMot}, with identical joint trajectories $q(t)$ (see Fig.~\ref{fig:dynTest_pos_equalizedSpecialInt}). 
	
	\subsubsection{Example~2 -- In reach end-effector task}
	\label{sec:endeffectortask}
	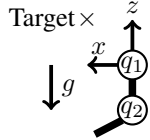
\begin{figure}[ht!]
		\centering
		\begin{tikzpicture}[scale=0.6]
		
		\node[circle,draw=black,inner sep=0pt, outer sep=0pt,thick,fill=white] (joint0) at ($(0,0)$) [] {$q_1$};
		\node[circle,draw=black,inner sep=0pt, outer sep=0pt,thick,fill=white] (joint1) at ($(0,-1)$) [] {$q_2$};
		
		\draw[line width=0.1cm] ($(joint0.south) + (0,0)$) -- ($(joint1.north)$);
		
		\draw[line width=0.1cm] ($(joint1.south west)$) -- ($(-0.83,-1.54)$);
		
		\draw[->,line width = 0.05cm] ($(-1.8,0)$) -- ($(-1.8,-1)$) node [midway,right] {${g}$};
		
		\begin{pgfonlayer}{bg} 
		\draw[->,line width = 0.05cm] ($(joint0)$) -- ($(joint0) + (0,1)$) node [midway,above=0.25] {${z}$};
		\draw[->,line width = 0.05cm] ($(joint0)$) -- ($(joint0) + (-1,0)$) node [near end=10,above] {${x}$};
		\end{pgfonlayer} 
		
		\node[] (target) at ($(-1,1)$) [] {$\times$};
		\node[] (targetdesc) at ($(-2.0,1)$) [] {Target};
		\end{tikzpicture}
		\caption[{Example~2, initial configuration of the robot.}]{\textbf{Example~2}, initial configuration of the robot. The desired end-effector position is at the cross.}
		\label{fig:2linkrobotex2}
	\end{figure}
	\begin{figure}[t!]
		\begin{tcolorbox}[title=Hierarchy for example 2,boxsep=1pt,left=2pt,right=3pt,top=2pt,bottom=1pt]
			\begin{enumerate}
				\setcounter{enumi}{0}
				\itemsep0em
				\item[L.$1$] Equation of motion, acceleration-based or velocity-based
				\item[L.$2$] Joint velocity limit (corresponds to the trust region constraint), expressed with accelerations by using finite differences~\eqref{eq:AccToVelForwardDifference} in the case of acc. based control 
				\item[L.$3$] {End-effector task with ${\ddot{e}}^{\text{ctrl}}$ (acc. based) or with ${\dot{e}}^{\text{ctrl}}_{\text{PD}}$ (vel. based) }
				\item[L.$4$] ${\dot{q}} = {0}$ (vel. based), expressed with accelerations by using forward differences~\eqref{eq:AccToVelForwardDifference} in case of acceleration-based control
				\item[L.$5$] ${\tau} = {0}$
			\end{enumerate}
		\end{tcolorbox}
		\caption{\textbf{Example~2}. Control hierarchy.}
		\label{fig:example2hierarchy}
	\end{figure}
	\begin{figure}[t!]
		\centering{\includegraphics[width=1\columnwidth]{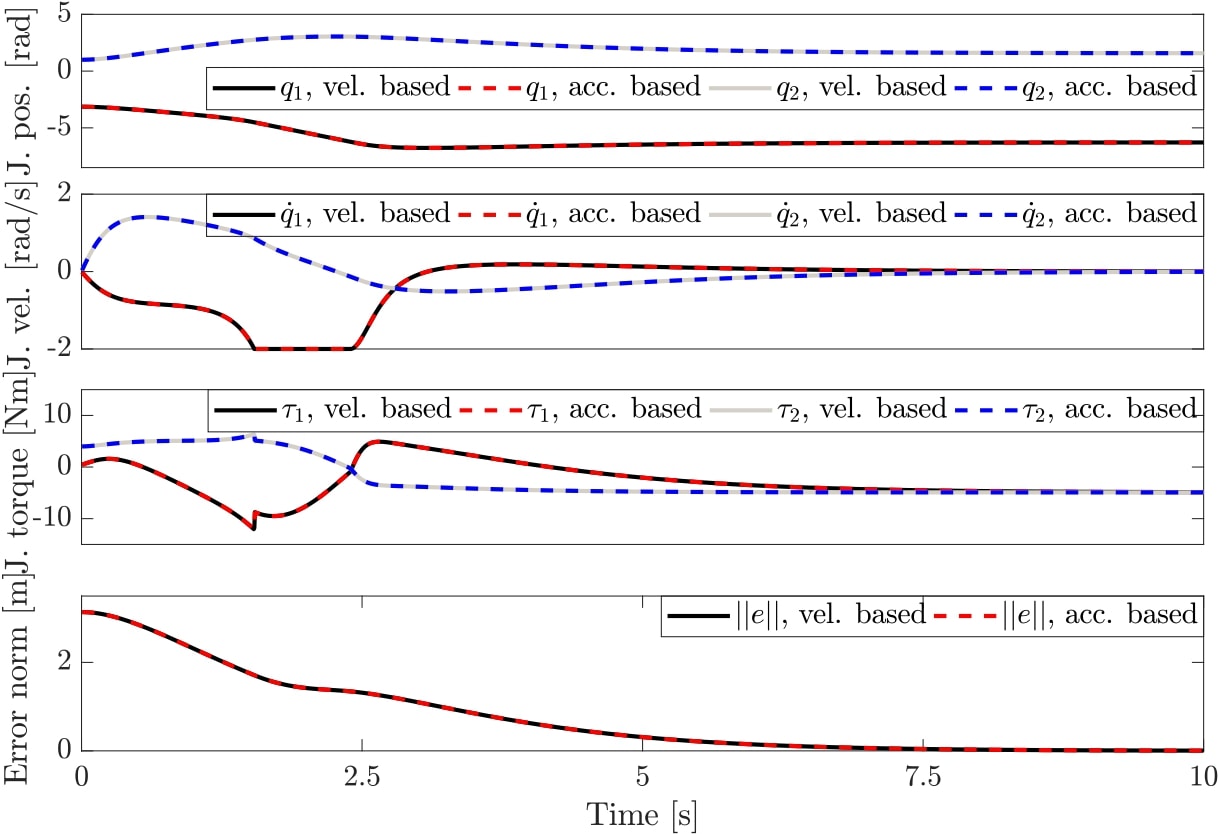}}
		\caption[Example 2, joint positions, joint velocities, joint torques and task error norm.]{\textbf{Example~2}, joint (J.) positions, velocities, torques and task error norm. They are identical for the acceleration- and velocity-based equation of motion and PD motion controllers ${\ddot{e}}^{\text{ctrl}}$ and ${\dot{e}}_{\text{PD}}^{\text{ctrl}}$.}
		\label{fig:dynTest3_2link}
	\end{figure}
	\begin{figure}[t!]
		\fboxsep=1mm%padding thickness
		\fboxrule=1pt%border thickness
		\centering{\fcolorbox{gray}{gray}{\includegraphics[width=0.5\columnwidth]{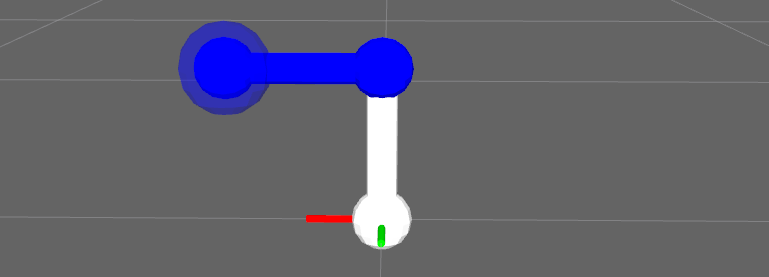}}}
		\caption[Example 2, converged robot posture.]{\textbf{Example~2}, converged robot posture.}
		\label{fig:dynTest3_convergedvel}
	\end{figure}
	
	steers the robot from the previous Example~1 to reach the desired Cartesian task-point $[1,1]$~m. The initial robot posture corresponds to an arbitrary non-singular configuration, e.g., $[-\pi,1]$~rad, see Fig.~\ref{fig:2linkrobotex2}. We define the acceleration- or velocity-based hierarchy in Fig.~\ref{fig:example2hierarchy}.
	
	The problem is solved by the GN algorithm (\eqref{eq:Ncontrol} with $\hat{H}_l = 0$). The joint positions, joint velocities, joint torques and task error norms are given in Fig.~\ref{fig:dynTest3_2link}. It can be seen that both the velocity and acceleration based control lead to an identical behavior. The converged robot posture is given in Fig.~\ref{fig:dynTest3_convergedvel}.
	
	\subsubsection{Example~3 -- Conflict between linearized task constraint and dynamics constraint}
	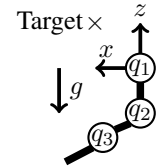
\begin{figure}[t!]
		\centering
		\begin{tikzpicture}[scale=0.6]
		
		\node[circle,draw=black,inner sep=0pt, outer sep=0pt,thick,fill=white] (joint0) at ($(0,0)$) [] {$q_1$};
		\node[circle,draw=black,inner sep=0pt, outer sep=0pt,thick,fill=white] (joint1) at ($(0,-1)$) [] {$q_2$};
		\node[circle,draw=black,inner sep=0pt, outer sep=0pt,thick,fill=white] (joint2) at ($(-0.83,-1.54)$) [] {$q_3$};
		\draw[line width=0.1cm] ($(joint0.south) + (0,0)$) -- ($(joint1.north)$);
		
		\draw[line width=0.1cm] ($(joint1.south west)$) -- ($(joint2.north east) + (0.05,-0.05)$);
		\draw[line width=0.1cm] ($(joint2.south west)$) -- ($(-1.66,-2.08)$);      
		\draw[->,line width = 0.05cm] ($(-1.8,0)$) -- ($(-1.8,-1)$) node [midway,right] {${g}$};
		
		\begin{pgfonlayer}{bg} 
		\draw[->,line width = 0.05cm] ($(joint0)$) -- ($(joint0) + (0,1)$) node [midway,above=0.25] {${z}$};
		\draw[->,line width = 0.05cm] ($(joint0)$) -- ($(joint0) - (1,0)$) node [near end=10,above] {${x}$};
		\end{pgfonlayer} 
		
		\node[] (target) at ($(-1,1)$) [] {$\times$};
		\node[] (targetdesc) at ($(-2.0,1)$) [] {Target};
		\end{tikzpicture}
		\caption[Example~3, initial configuration of the robot.]{\textbf{Example 3}, initial configuration of the robot. The desired end-effector position is at the cross.}
		\label{fig:3linkrobot}
	\end{figure}
	\begin{figure}[t!]
		\begin{tcolorbox}[title=Hierarchy for example 3,boxsep=1pt,left=2pt,right=3pt,top=2pt,bottom=1pt]
			\begin{enumerate}
				\setcounter{enumi}{0}
				\itemsep0em
				\item[L.1] 
				\begin{itemize}
					\itemsep0em
					\item Equation of motion, acceleration-based or velocity-based with accelerations integrated by~\eqref{eq:AccToVelForwardDifference}
					\item $\tau_1 = 0$
				\end{itemize}
				\item[L.2] Trust region constraint,  expressed with accelerations by using forward differences~\eqref{eq:AccToVelForwardDifference} in the case of acc. based control 
				\item[L.3] \textbf{End-effector task with ${\ddot{e}}^{\text{ctrl}}$ (acc. based) or with ${\dot{e}}^{\text{ctrl}}_{\text{PD}}$ (vel. based), solved by the GN algorithm, Newton's method (vel. based) or the LM algorithm (acc. based)}
				\item[L.4] ${\dot{q}} = {0}$ (vel. based), expressed with accelerations by using forward differences~\eqref{eq:AccToVelForwardDifference} in case of acceleration-based control
				\item[L.5] ${\tau} = {0}$
			\end{enumerate}
		\end{tcolorbox}
		\caption{\textbf{Example~3}. Control hierarchy.}
		\label{fig:example3hierarchy}
	\end{figure}
	\begin{figure}[t!]
		\centering{\includegraphics[width=1\columnwidth]{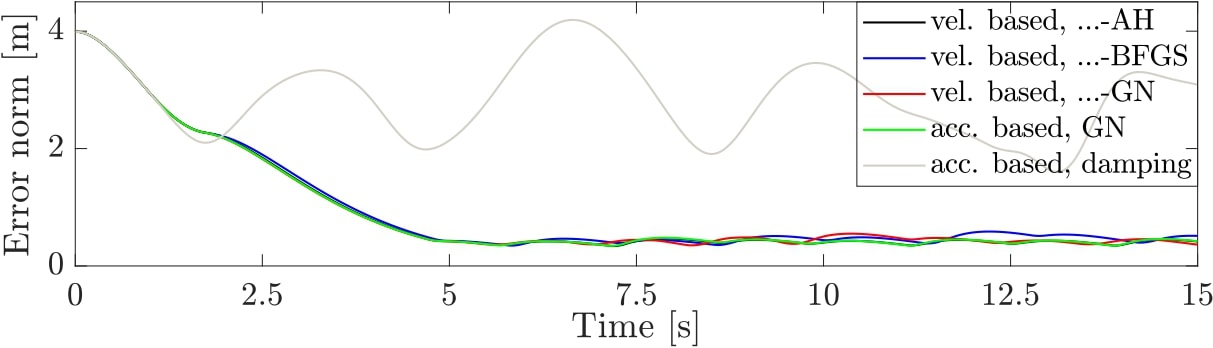}}
		\caption[Example~3, task error norm for the end-effector task.]{\textbf{Example 3}, task error norm for the end-effector task. Damping in the acceleration domain leads to overshooting behavior and the error is barely minimized (dark gray curve). The velocity-based PD controller ${\dot{e}}_{\text{PD}}^{\text{ctrl}}$ prevents this behavior.}
		\label{fig:dynTest2_errorNorm}
	\end{figure}
	
	%\begin{figure}[t!]
	%	\centering{\includegraphics[width=1\columnwidth]{figures/dynTest2_tau.jpg}}
	%	\caption[Example 3, joint torques.]{\textbf{Example 3}, joint torques. The acc. and vel. based GN algorithm leads to numerical instabilities in the joint torques of joint 2 and 3. LexDynAH and LexDynBFGS are numerically stable.}
	%	\label{fig:dynTest2_tau}
	%\end{figure}
	\begin{figure}[t!]
		\centering{\includegraphics[width=1\columnwidth]{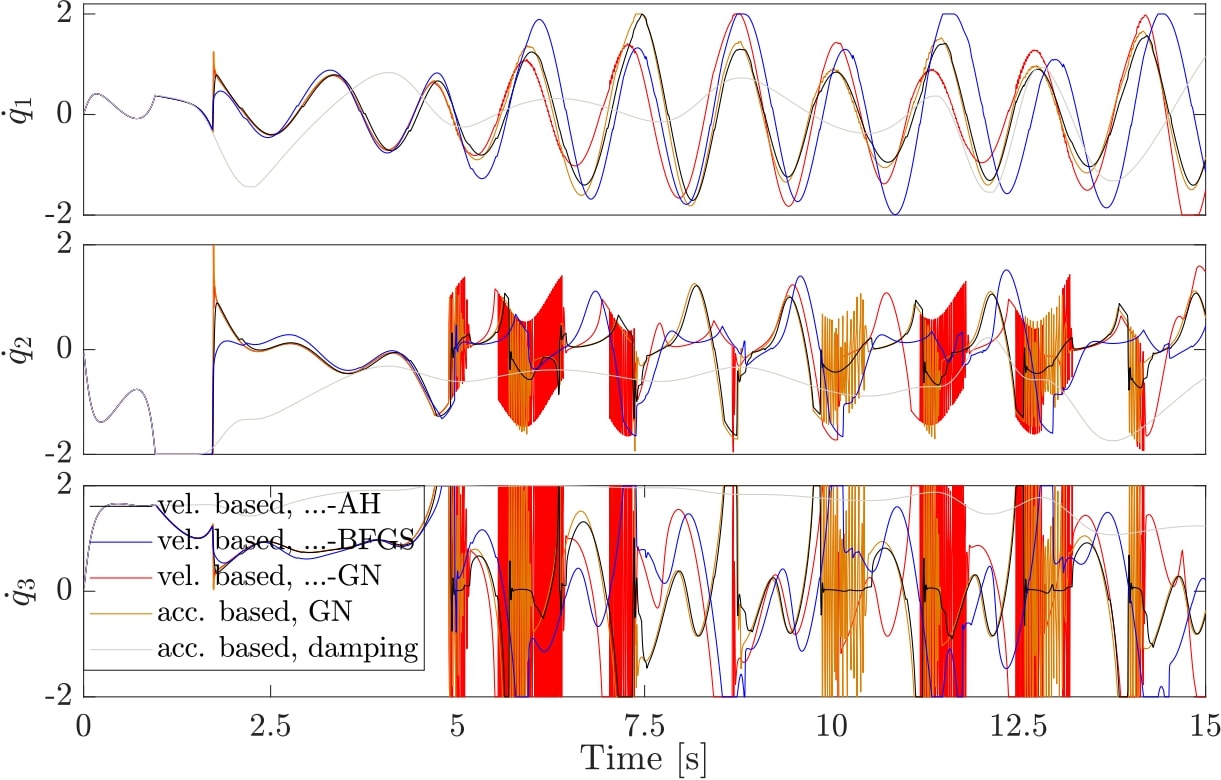}}
		\caption[Example 3, joint velocities.]{\textbf{Example 3}, joint velocities. The acc. and vel. based GN algorithm leads to numerical instabilities in the joint velocities $\dot{q}_2$ and $\dot{q}_3$. They occur whenever the kinematic subchain consisting of link 2 and link 3 is close to kinematic singularity. LexDynAH (AH) and LexDynBFGS (BFGS) are numerically stable. Damping in the acceleration domain is numerically stable but leads to overshooting behavior.}
		\label{fig:dynTest2_vel}
	\end{figure}
	\begin{figure}[t!]
		\fboxsep=1mm%padding thickness
		\fboxrule=1pt%border thickness
		\centering{\fcolorbox{gray}{gray}{\includegraphics[width=0.5\columnwidth]{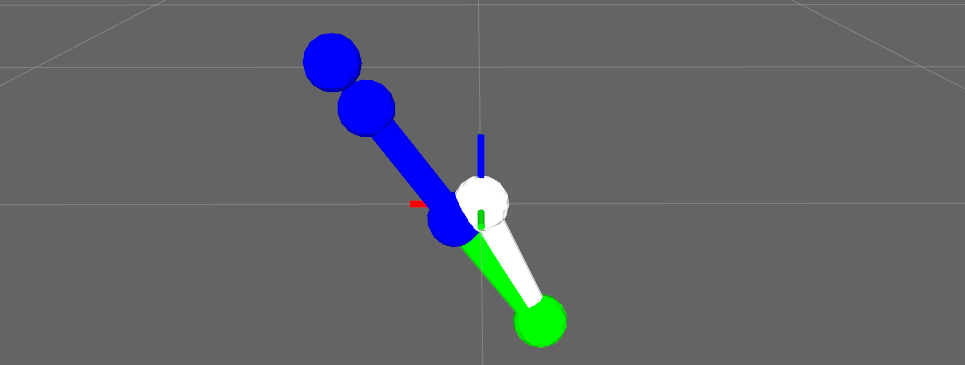}}}
		\caption[Example 3. Converged robot posture.]{\textbf{Example 3}. The robot managed to reach the target as close as possible (LexDynAH). However, some swaying motion remains due to the freely swinging joint 1 (white ball) with zero commanded torque. Note how the kinematic subchain of the green and blue link is in kinematic singularity.}
		\label{fig:dynTest2_convergedvel}
	\end{figure}
	
	In this simulation we want to test the behavior of the robot if a linearized end-effector task is in conflict with a (non-linearized) dynamics constraint. 
	%Numerical instabilities due to algorithmic singularities with higher level non-linearized tasks only occur if the higher priority task forces kinematic subchains of the lower priority linearized end-effector task into kinematic singularity. Therefore, 
	We use a robot with 3-DoF's. We set its initial configuration to $[-\pi,1,0]$~rad, see Fig.~\ref{fig:3linkrobot}.
	The hierarchy is given in Fig.~\ref{fig:example3hierarchy}. Constraints in bold have to be considered in the calculation of the hierarchical Hessian. Note that for bound constraints we have ${H} = {0}$.
	The commanded joint~1 torque is set to zero and hence conflicts with the end-effector task of reaching the point $[1,1]$~m. 
	The trust region constraint is put on the second level in order to not interfere with the equation of motion.
	
	The norm of the distance of the end-effector from the target is given in Fig.~\ref{fig:dynTest2_errorNorm}. As discussed in Sec.~\ref{sec:dampAcc}, damping in the acceleration domain leads to slowly oscillating behavior that can be observed from the error curve `acc. based, damping'. The damping value is set to $\mu = 0.1$. The error norm of the other methods gets minimized fairly well but is disturbed by the slow oscillations of the freely swinging joint~1. 
	
	Figure~\ref{fig:dynTest2_vel} shows the joint velocities of the robot. For the GN algorithm (both acc. and vel. based) numerical instabilities can be observed. This is due to the zero commanded joint~1 torque that conflicts with the aim of getting closer to the target with the end-effector. In order to minimize the task error as much as possible the kinematic subchain consisting of both links~2 and~3 is forced into kinematic singularity.
	
	Newton's method with second order information from both LexDynAH and LexDynBFGS leads to numerically stable joint velocity behavior without overshooting the end-effector as seen for the damping in acceleration-based control. The robot configuration with low error norm but with remaining some swaying motion is given in Fig.~\ref{fig:dynTest2_convergedvel}.
	
	\subsection{Experiments with a humanoid robot}
	\label{sec:realTests}
	
	\begin{figure}
		\begin{tcolorbox}[title=Hierarchy for LexDynAH and LexDynBFGS,boxsep=1pt,left=2pt,right=3pt,top=1pt,bottom=1pt]
			\begin{enumerate}
				\setcounter{enumi}{0}
				\itemsep0em
				\setlength{\itemindent}{1.5mm}
				\item[L.1]
				\begin{itemize}
					\itemsep0em
					\item $4n_c$ ($n_c$: number of contacts) bounds on generalized contact wrenches: ${\gamma} > {0}$
					\item $32$ joint limits using a velocity damper~\cite{faverjon1987}
				\end{itemize}
				\item[L.2]
				\begin{itemize}
					\itemsep0em
					\item $38$ integrated equations of motion
					\item $38$ torque limits
				\end{itemize}
				\item[L.3] $38$ trust region limits
				\item[L.4] \textbf{$3$ inequality constraints for self-collision avoidance}
				\item[L.5] \textbf{$18$ geometric contact constraints}
				\item[L.6] $1$ equality constraint on the head yaw joint to put the vision marker into the field of view
				\item[L.7] \textbf{$3$ inequality constraints on the CoM}
				\item[L.8]
				\begin{itemize}
					\itemsep0em
					\item \textbf{$6$ end-effector equality constraints for left and right hand}
					\item \textbf{$3$ equality constraints to keep the chest orientation upright}
				\end{itemize}
				\item[L.9] \textbf{$3$ stricter inequality constraints on the CoM}
				\item[L.10] $38$ constraints to minimize joint velocities: ${\dot{q}} = {0}$
				\item[L.11] ${4n_c}$ constraints to minimize the generalized contact wrenches: ${\gamma} = {0}$
			\end{enumerate}
		\end{tcolorbox}
		\caption{Control hierarchy for HRP-2Kai experiments.}
		\label{fig:hrp2hierarchy}
	\end{figure}
	
	Here, we assess our approach with a real humanoid robot HRP-2Kai. We devised a control hierarchy given in Fig.~\ref{fig:hrp2hierarchy}.
	The hierarchical separation of the bound constraints on the generalized contact wrenches and the joint limits ($l=1$) from the equation of motion ($l=2$) allows LexLSI to cheaply handle the variable bounds on the first level.

	\begin{figure}
		\begin{tcolorbox}[title=Hierarchy for WLS,boxsep=1pt,left=2pt,right=3pt,top=2pt,bottom=1pt]
			\begin{enumerate}
				\setcounter{enumi}{0}
				\itemsep0em
				\item[L.1]
				\begin{itemize}
					\itemsep0em
					\item $4n_c$ bounds on generalized contact wrenches: ${\gamma} > {0}$
					\item $32$ joint limits using an acceleration damper~\cite{faverjon1987}
					\item $38$ equations of motion
					\item $38$ torque limits
					\item $3$ inequality constraints for self collision avoidance
					\item $18$ geometric contact constraints
					\item $3$ inequality constraints on the CoM
				\end{itemize}
				\item[L.2]
				\begin{itemize}
					\itemsep0em
					\item $6$ end-effector equality constraints for left and right hand
					\item $32$ equality constraints to maintain a reference posture (with the head yaw joint turned towards the vision marker)
					\item $4n_c$ constraints to minimize the generalized contact wrenches:
					$
					{\gamma} = {0}
					$
				\end{itemize}
			\end{enumerate}
		\end{tcolorbox}
		\caption{Weighted least-squares control hierarchy}
		\label{fig:hierarchywls}
	\end{figure}
	
	For comparison purpose, we use the acceleration-based weighted control hierarchy given in Fig.~\ref{fig:hierarchywls}, and solved by LSSOL.
	The hierarchy for WLS contains the dynamic constraints (equation of motion and bounds on ${\gamma}$ and ${\tau}$), joint limits, self-collision avoidance, center of mass (CoM) task and contact tasks as constraints on priority level~$1$.
	All these tasks have the same priority without weighting. Although the self-collision avoidance, the contact and the CoM constraints can be the source of potential conflict or even (unresolved) kinematic singularities, there are no regularization tasks in order to avoid interaction with the equation of motion. This highlights the significance of being able to easily design safe hierarchical robot control problems with LexDynAH or LexDynBFGS.
	The reaching task is defined as an objective on priority level~$2$.
	A posture reference task is also added at the objective level. Similarly to the LM algorithm, it acts as a velocity damper to approach singular configurations; it has a low weight ($5 \cdot 10^{-2}$) that is tuned depending on the task to be performed. For example reaching for very far away targets requires a higher weight.
	A further regularization task ${\gamma} = {0}$ on the contact forces is added to yield a fully determined and full-rank problem. It is weighted with a small value $10^{-4}$.
	
	For both solvers we substitute the torques ${\tau}$ with the equation of motion which reduces the number of variables from $112$ to $74$.
	
	\subsubsection{Experiment~1}
	\begin{figure}[t!]
		\centering{\includegraphics[width=1.\columnwidth]{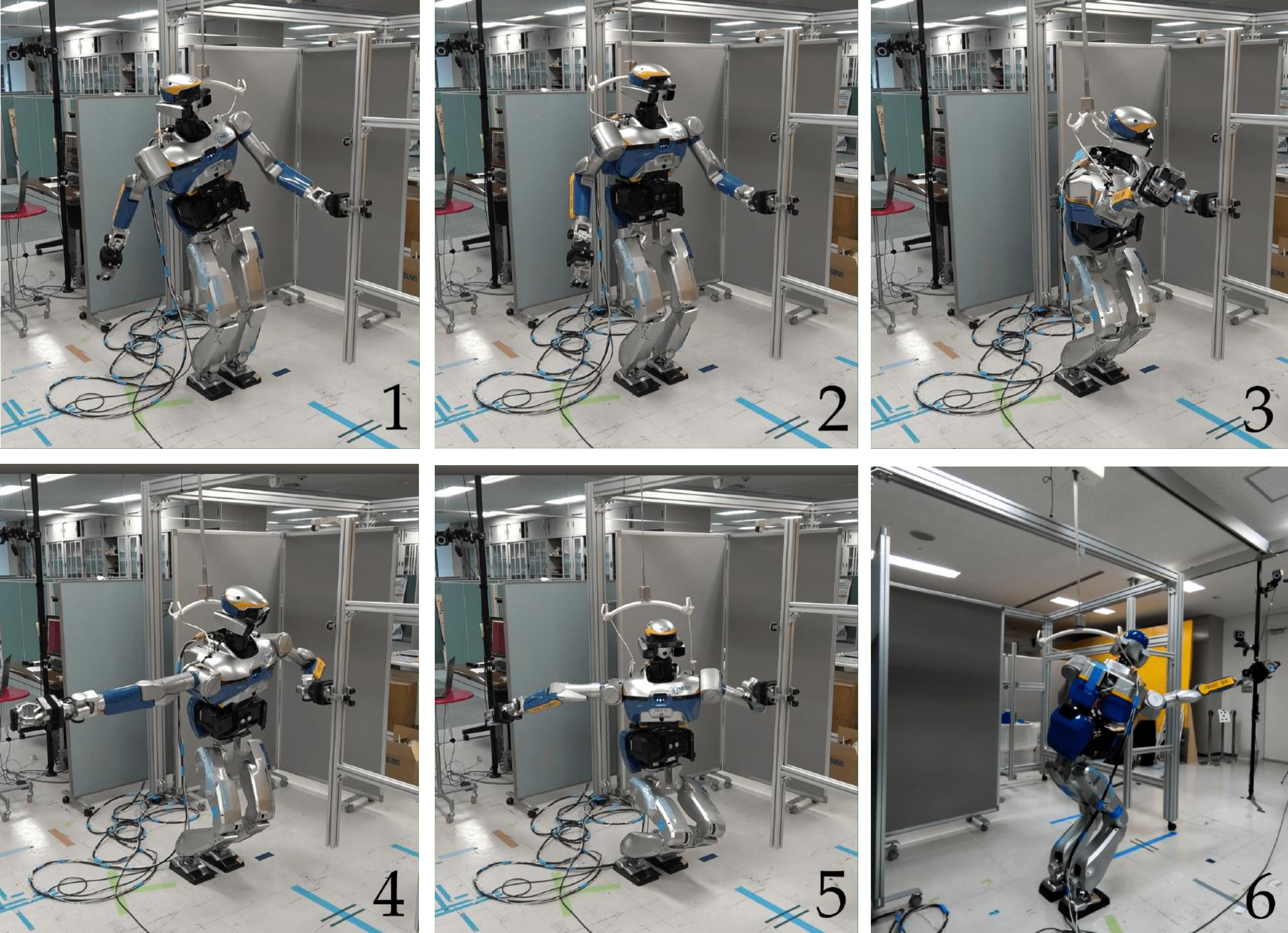}}
		\caption{Pictures of HRP-2Kai performing \textbf{exp. 1}, LexDynAH, from left to right: \textbf{1.:} The left hand has grabbed the pole while the right hand task on level 9 is in conflict with the CoM task on level 8. \textbf{2.:} The CoM on level 8 switched from box~1 to box~2 and is not in conflict with the right hand task anymore. The right hand task on level 9 is not augmented anymore. \textbf{3.:} HRP-2Kai is in full forward stretch. \textbf{4.:} The robot moves its right hand to the back. \textbf{5.:} HRP-2Kai is in full backward stretch. \textbf{6.:} The robot during its second forward stretch.}
		\label{fig:exp2_screenshots}
	\end{figure}
	% LexDynBFGS
		\begin{figure}[t!]
		\centering{\includegraphics[width=1\columnwidth]{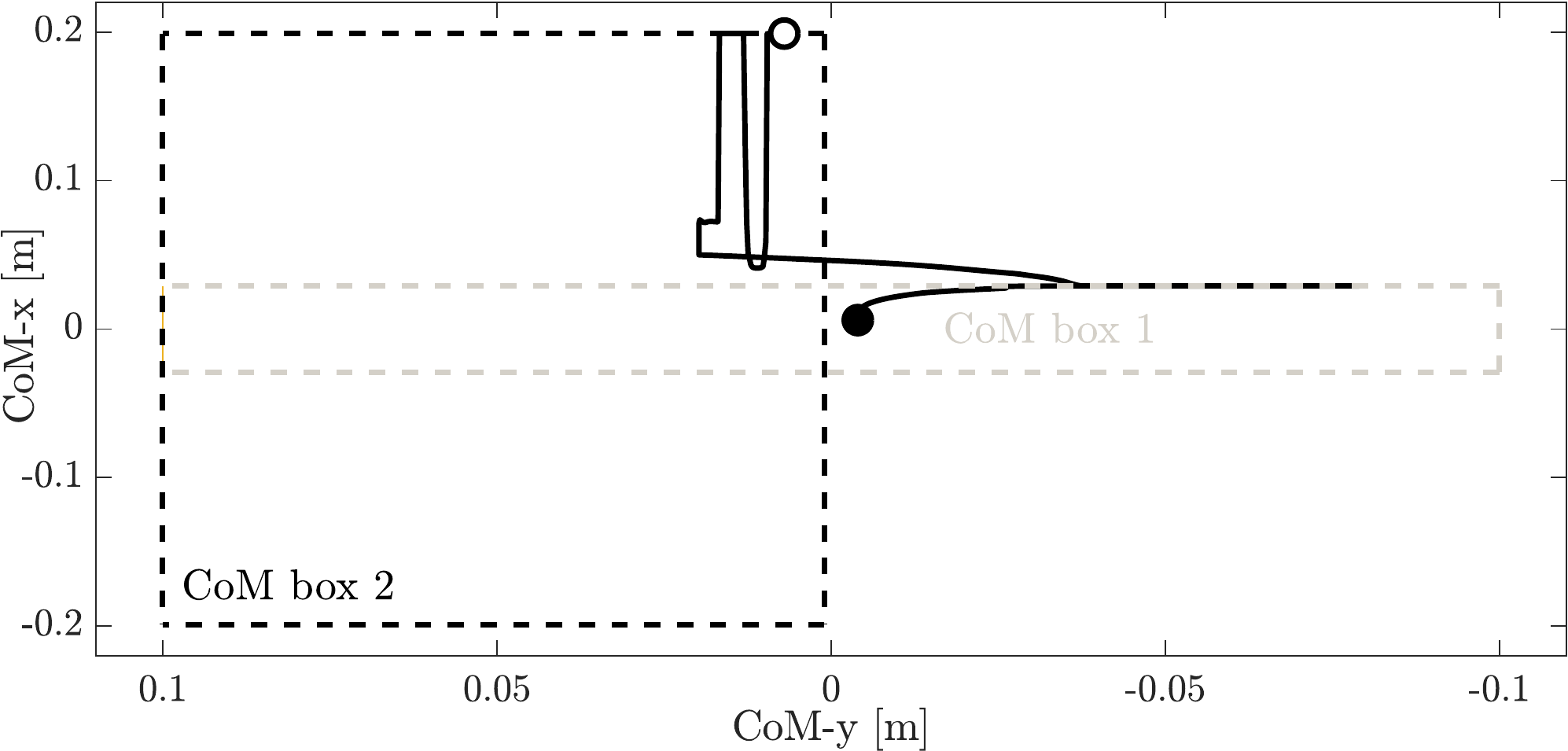}}
		\caption{\textbf{Exp. 1}, LexDynAH, CoM movement. The CoM starts at the black dot and moves until it arrives at the white dot, first being constrained in box 1 and then in box 2.}
		\label{fig:exp2_LexDynAH_com}
	\end{figure}
	\begin{figure}[t!]
	\centering{\includegraphics[width=1\columnwidth]{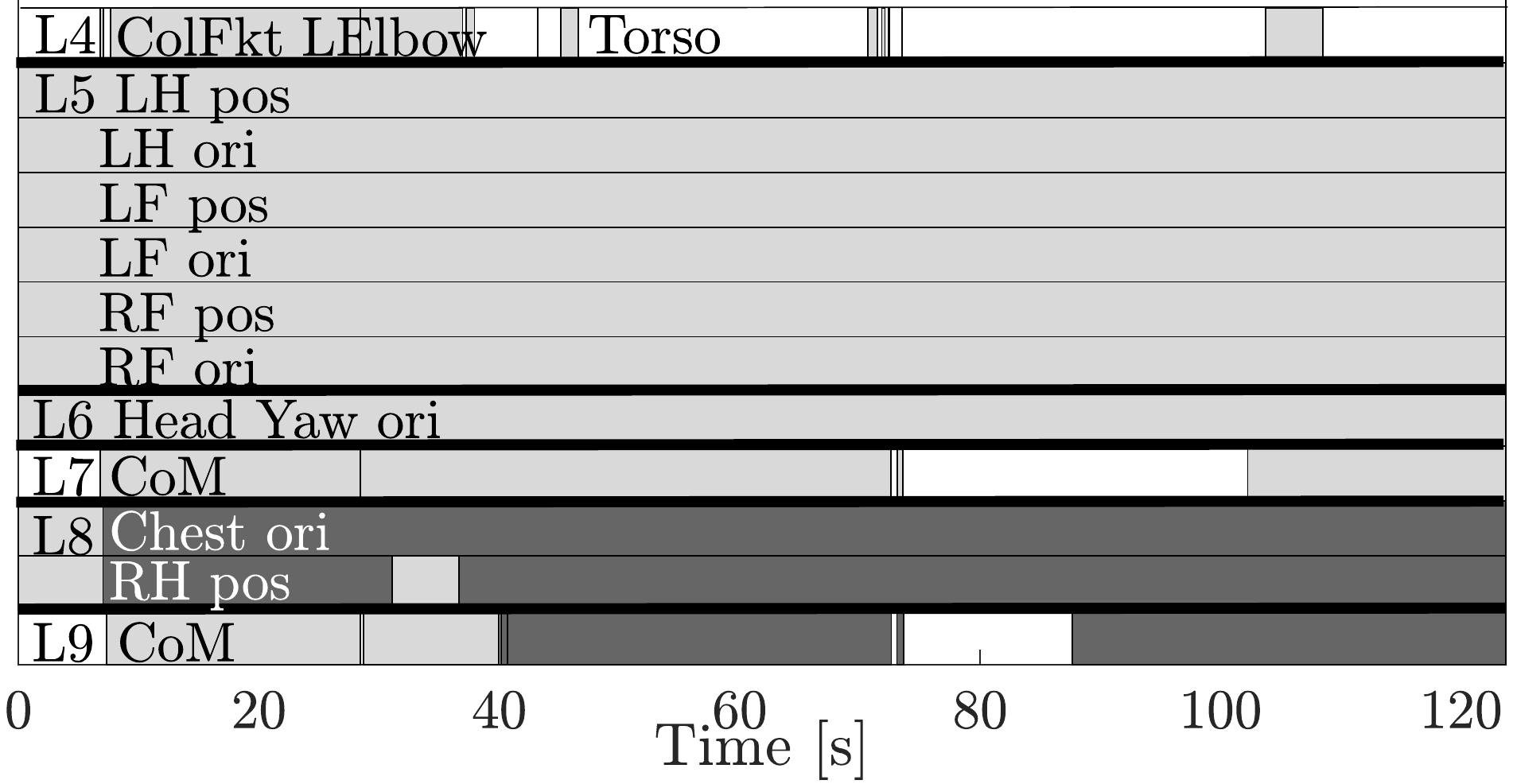}}
	\caption{\textbf{Exp. 1}, LexDynAH, map of activity (light gray) and Newton's method (dark gray)}
	\label{fig:exp2_LexDynAH_HOnASMap}
\end{figure}
	\begin{figure}[t!]
	\centering{\includegraphics[width=1\columnwidth]{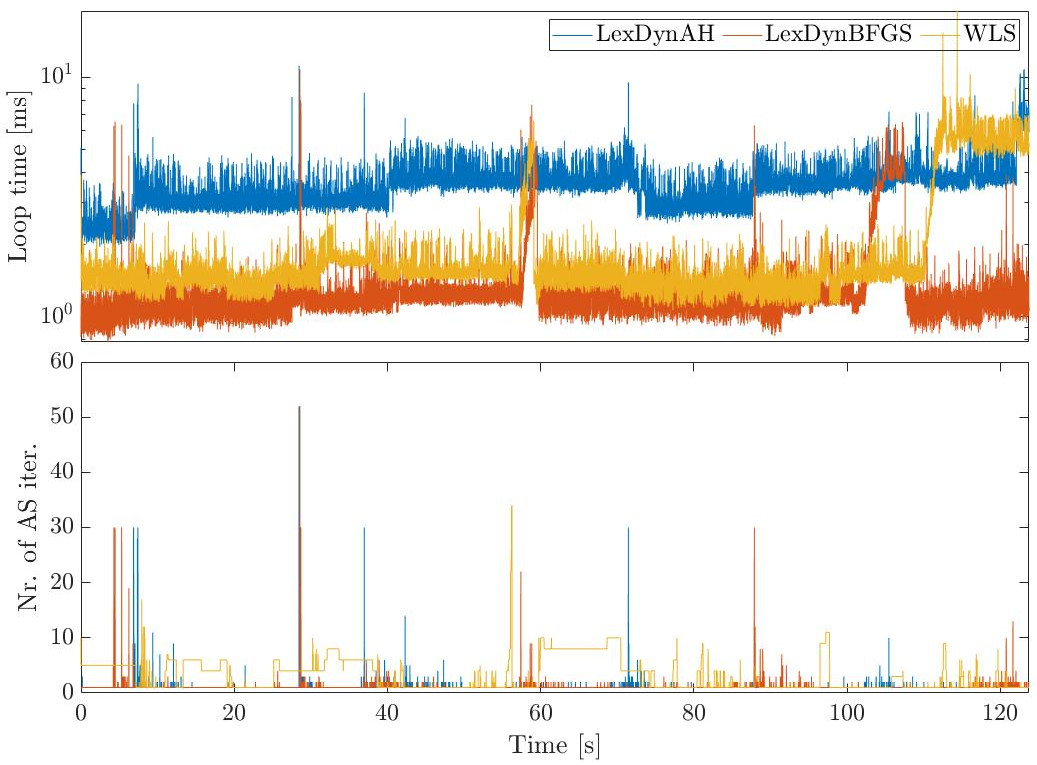}}
	\caption{\textbf{Exp. 1}, computation times and active set iterations. LexDynAH peaks at 28 s with 2.63 ms and a maximum number of active set iterations of 52. LexDynBFGS peaks at 28 s with 1.11 ms and a maximum number of active set iterations of 52. WLS peaks at 56 s with 2.94 ms and a maximum number of active set iterations of 34.}
	\label{fig:exp2_LexDynAH_LoopTimes}
\end{figure}
	\begin{figure}[t!]
		\centering{\includegraphics[width=1\columnwidth]{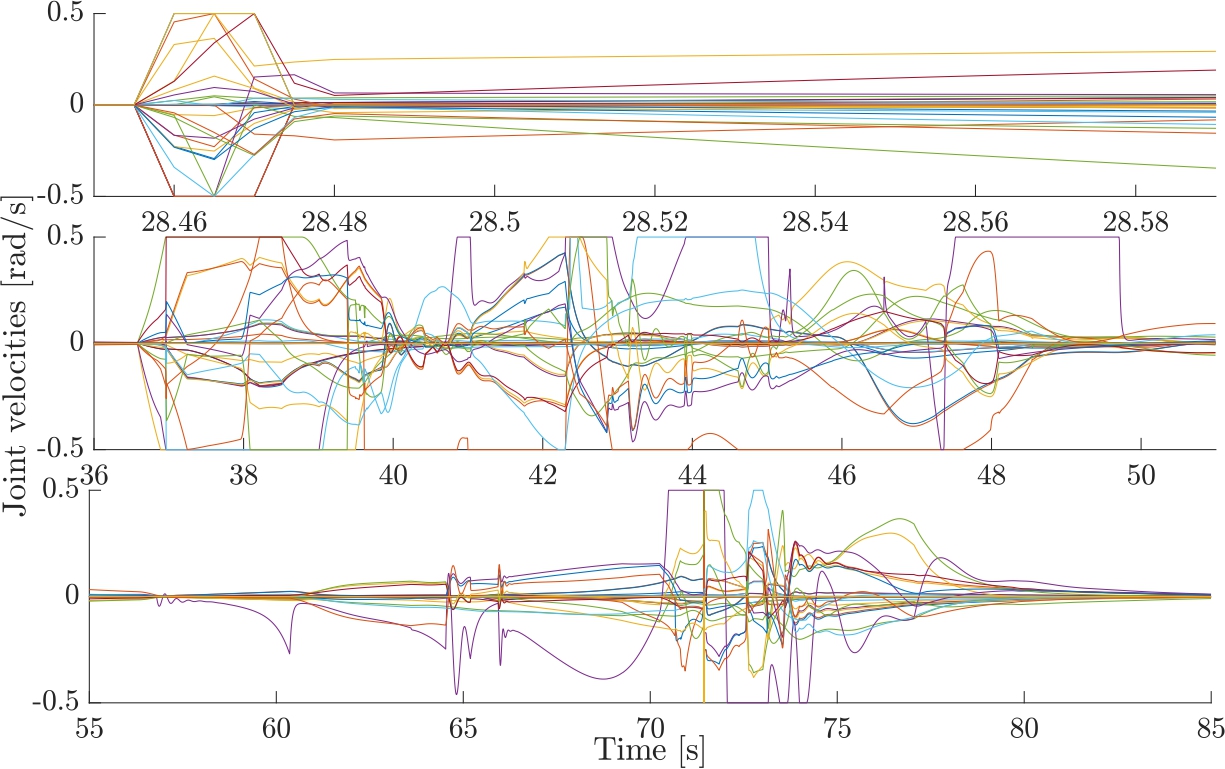}}
		\caption{\textbf{Exp. 1}, LexDynAH, joint velocities. The upper graph shows the moment when the CoM is released and the 52 active set iterations occur (followed by 5 occurrences of 30 iterations until 32.7 s). The middle one shows the moment when the right hand stretches to the right and starts being augmented. The lower graph shows the joint velocities when the robot stretches to the back and crouches.}
		\label{fig:exp2_LexDynAH_JointVel}
	\end{figure}
	\begin{figure}[t!]
	\centering{\includegraphics[width=1\columnwidth]{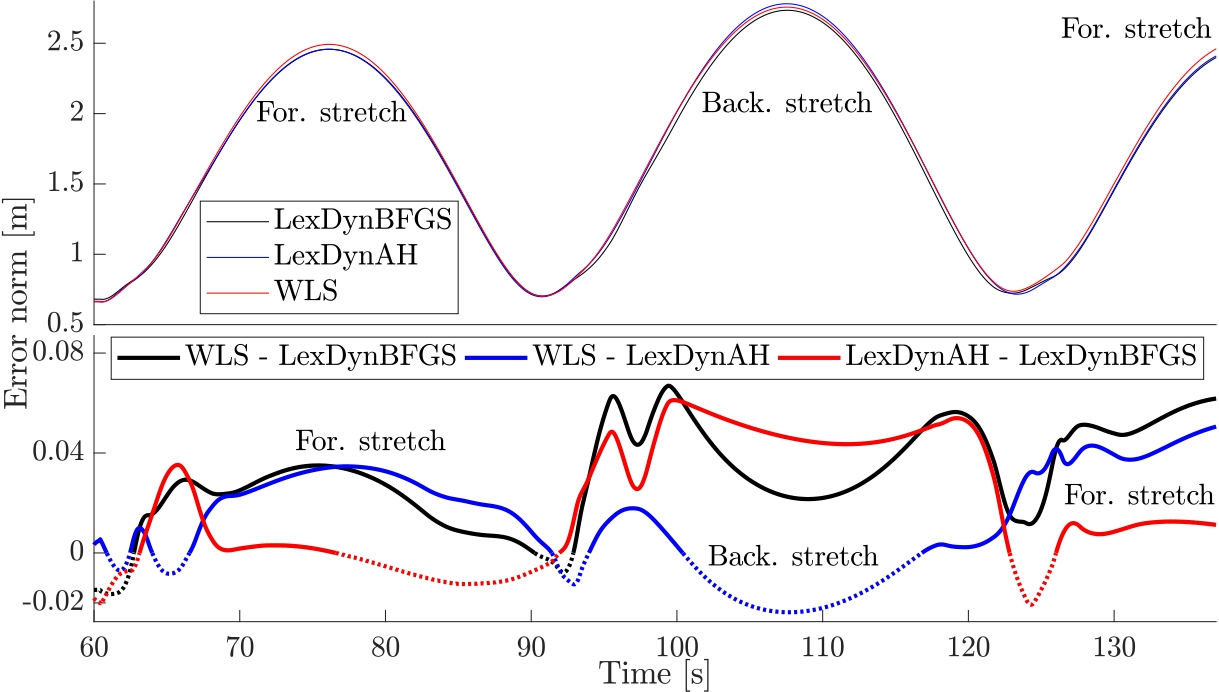}}
	\caption{\textbf{Exp. 1}, comparison of the error norm of the right hand tracking the swinging target during forward (for.) and backward (back.) stretch. The lower graph shows the differences of the error norms of the different methods.}
	\label{fig:exp2_errorComp}
\end{figure}
	
	% WLS
	\begin{figure}[t!]
		\centering{\includegraphics[width=1\columnwidth]{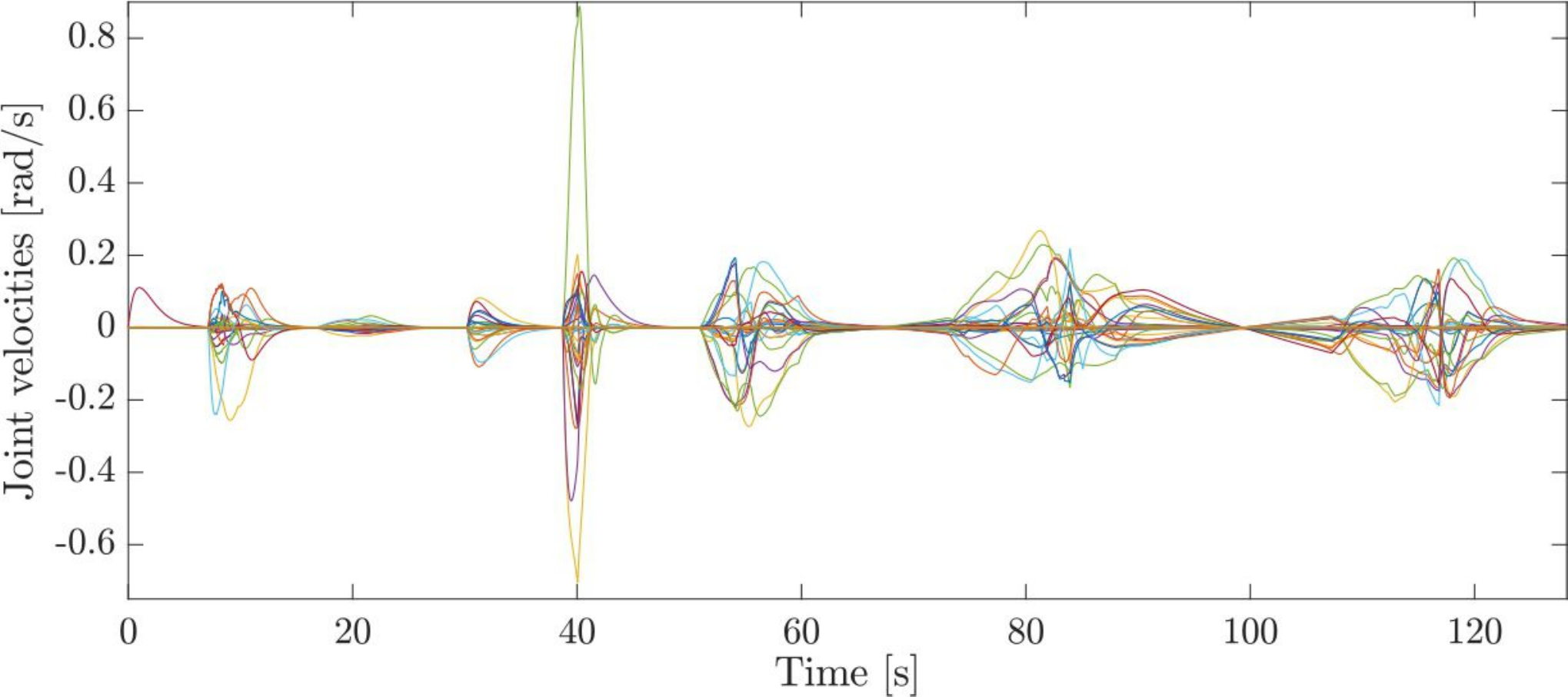}}
		\caption{\textbf{Exp. 1}, WLS, joint velocities.}
		\label{fig:exp2_wls_JointVel}
	\end{figure}
	
	The first experiment (exp.~1, see Fig.~\ref{fig:exp2_screenshots}) is designed in such a way that a third contact between the left hand and a rigid pole must be established in order to prevent falling. This is a fixed bilateral contact that allows pulling or pushing forces.
	
	In the following we describe the experiment for LexDynAH. 
	At first, the robot moves its left hand to the vertical metal pole and grabs it. The pole position is determined by vision thanks to a marker provided by the {\tt whycon} library~\cite{nitsche:iros:2015}. During the movement, the CoM task on level~$7$ is constrained by the bounding box~1 $[\pm 0.03, \pm 0.1, \pm\infty]$~m and gets activated, see Fig.~\ref{fig:exp2_LexDynAH_com}. The right hand on the next level~$8$, which must remain at its current position, is therefore in conflict and moves slightly backwards. This is a purely algorithmic singularity and triggers the switch to Newton's method. Figure~\ref{fig:exp2_LexDynAH_HOnASMap} shows the activation of the CoM task and the augmentation of the right hand position task and the chest orientation task at around $8$~s.
	
	Box~1 is widened from $[\pm 0.03, \pm 0.1, \pm\infty]$~m to $[\pm 0.2, 0.05 \pm 0.05, \pm\infty]$~m since we increase the support area of the robot in the sagittal plane due to the additional contact. However, several control iterations with a high number of active set is made to adjust the robot state; in Fig.~\ref{fig:exp2_LexDynAH_LoopTimes} at $28$~s notice the active set iterations peak ($52$). Figure~\ref{fig:exp2_LexDynAH_HOnASMap} shows that at the instant of CoM release (around $28$~s), both CoM tasks at level~7 and~9 are activated and shortly after deactivated again. Similar behavior is seen for the collision constraint between the left elbow and the torso.
	The upper graph of Fig.~\ref{fig:exp2_LexDynAH_JointVel} shows how the velocity suddenly increases from zero to maximum velocity $0.5$~s during the CoM release. The corresponding dynamic effects are also adjusted (the trust region and the contact force constraints, with possible interplay), resulting in a large number of active set iterations.
	
	At around $36$~s the augmentation of the level~$8$ position task stops as the right hand reached its prescribed position, see Fig.~\ref{fig:exp2_LexDynAH_HOnASMap}.
	The right hand then follows a target swinging from the front $[3, -1.5,2]$~m to the back $[-3, -1.5,0]$~m of the robot (for reference, HRP-2Kai is $1.71$~m tall with a $2.11$~m arm span). The target is always out of reach but numerically stable robot control is obtained thanks to the switch to the Newton's method. As can be seen from the middle and bottom graph of Fig.~\ref{fig:exp2_LexDynAH_JointVel}, the joint velocities are numerically stable and do not show signs of high frequency oscillations as for example observed in Fig.~\ref{fig:dynTest2_vel} for the GN algorithm. The joint velocities thereby exhibit a bang-bang type of behavior (generating joint velocities at the limits) for as fast as possible task convergence~\cite{pham2013}. 
	During the stretch motions, the CoM is well outside the support polygon of the feet (approximately at $[\pm 0.1,\pm 0.2,\pm \infty]$~m) which stresses the importance of the left hand supporting contact (see Fig.~\ref{fig:exp2_LexDynAH_com}). 
	
	Figure~\ref{fig:exp2_errorComp} shows the norms of the tracking error of the right hand during the stretch motions.
	During forward stretches LexDynAH performs better than WLS ($\textrm{Err}_{\textrm{WLS}}-\textrm{Err}_{\textrm{LexDynAH}} > 0$).
	Especially during the second forward stretch (130s+) LexDynAH gets over $0.05$~m closer to the target than WLS due to fully stretching its arm leading to a kinematic singularity. 
	LexDynAH has a higher error norm than WLS during the backward stretch (see Fig.~\ref{fig:exp2_errorComp}, 100~s to 117~s). This is despite the fact that the robot crouches more than seen for WLS (see video). This can be explained by local minima of the non-linear kinematic optimization problem created by joint limits.
	
	Figure~\ref{fig:exp2_LexDynAH_LoopTimes} shows that the computation time of LexDynAH, which includes the calculation time of the analytic Hessian, increases by about $1.5$~ms as the right hand task gets augmented at time 8~s. An additional increase can be observed at $40$~s when the CoM task on level~9 also requires augmentation (see Fig.~\ref{fig:exp2_LexDynAH_HOnASMap}). Now two expensive symmetric Schur decompositions of the hierarchical Hessian on level~8 and~9 are required.
	
	LexDynBFGS is very similar to LexDynAH and shows the capabilities of the BFGS algorithm to provide a valid approximation of the analytic hierarchical Hessian. Computation times (Fig.~\ref{fig:exp2_LexDynAH_LoopTimes}) are lower than for LexDynAH since the hierarchical Hessian approximation is already positive definite.
	
	WLS exhibits very smooth joint velocities (see Fig.~\ref{fig:exp2_wls_JointVel}) due to the conservatively chosen damping factor. This comes at the cost of worse convergence, especially compared to LexDynBFGS, see Fig.~\ref{fig:exp2_errorComp}. Especially during the forward stretches the robot does not fully stretch its right arm. Also during the backward stretch the robot crouches to a lesser extent than seen for LexDynAH and LexDynBFGS (see video).

	\subsubsection{Experiment 2}
	
	\begin{figure}[t!]
		\centering{\includegraphics[width=1\columnwidth]{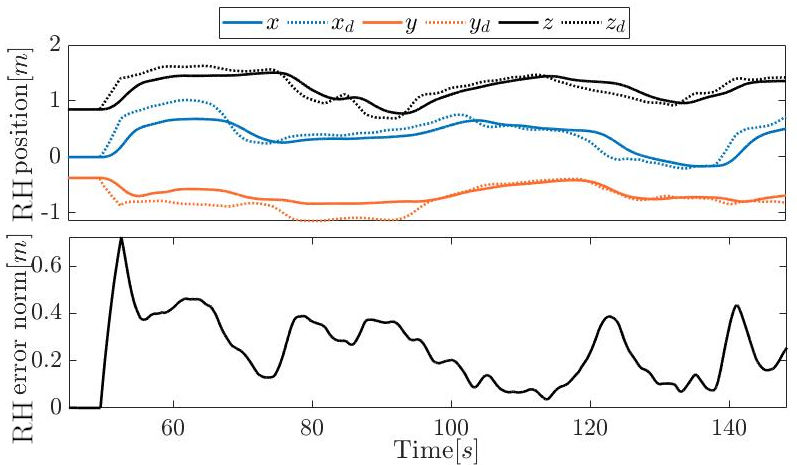}}
		\caption{\textbf{Exp.~2}, LexDynAH, right hand task error. The upper graph shows the actual right hand position in $x$, $y$ and $z$ direction and the desired ones $x_d$, $y_d$ and $z_d$. The lower graph shows the error norm.}
		\label{fig:exp3_LexDynBFGS_taskError}
	\end{figure}
	
	With the second experiment (exp.~2) we show the importance of handling kinematic and algorithmic singularities without the reliance on heuristic damping tuning.
	This experiment is set up similarly to exp.~1. This time however the robot's right hand tracks a target held by a human operator which is sampled by a motion capture system. In such situations, and especially for a humanoid robot with multi-level prioritized constraints, it is difficult to determine the robot workspace beforehand.
	Our proposed methods LexDynAH and LexDynBFGS allow to only have approximate knowledge or no knowledge at all of the robot workspace while still enabling safe control of the robot. The target could be at the border of the workspace with the end-effector being in kinematic singularity or being in algorithmic conflict with a stable-balance task like the CoM task on a higher priority level. At the same time, the whole possible workspace is used. This might be in contrast to a conservatively tuned regularization term in order to ensure non-singular behavior over a range of robot configurations which might not be known in advance.
	
	The right hand tracking error of LexDynAH is given in Fig.~\ref{fig:exp3_LexDynBFGS_taskError}. From around $50$~s onwards, the right hand is tracking the position of a handheld wand with markers detected by the motion capture system. The provided position $x_d$, $y_d$ and $z_d$ is filtered by a $3$~s moving average filter (see dotted lines in upper graph of Fig.~\ref{fig:exp3_LexDynBFGS_taskError}). Over all the course of the experiment the robot exhibits a numerically stable control. The marker is moved outside, on the border and inside of the robot's workspace with a left hand (LH) error norm of up to 70~cm. Note that in order to prevent too fast and unsafe motions, especially when the marker is far away, we set the right hand proportional task gain to $k_p = 0.5$ instead of $k_p = 1$ in the previous experiment. This leads to rather slow robot movements with relatively slow convergence behavior. 
	
	\subsection{Performance comparison}
	The overall performance of LexDynBFGS and LexDynAH in the experiments is compiled comprehensively in Table~\ref{table:performance}.
	
	\begin{table}[ht!]
		\centering
		\tabcolsep=0.185cm
		\begin{tabular}{l|cccr}
			& \textbf{LexDyn} & \textbf{LexDyn}  & \textbf{WLS}\\
			& \textbf{AH} & \textbf{BFGS}  &
			\\
			\hline
			Error convergence & + & + & o / -- \\
			Joint stability & + & + & + / -- \\
			Joint smoothness & o & o & + / --\\
			Easiness of use & + & + & o\\
		\end{tabular}
		\caption{Comprehensive overview of the performance of LexDynAH, LexDynBFGS and WLS in the evaluation. The symbols +, o, -- indicate best, acceptable and worst performance in the corresponding evaluation criteria.}
		\label{table:performance}
	\end{table}
	LexDynAH and LexDynBFGS both performed best in the category `Error convergence'. The performance of WLS depends highly on the chosen damping. Its behavior can only be considered acceptable at best since determining `perfect' damping can not be achieved with adaptive heuristics.
	The same holds for the `stability' and `smoothness' of the joint trajectories. If the damping weights for the posture reference task of WLS are too low, the numerical behavior is unstable and therefore not smooth. If the damping weights are too high we get very smooth joint trajectories but bad error convergence. LexDynAH and LexDynBFGS are numerically stable alternatives with relatively smooth joint trajectories. Especially, there is no need for damping tuning which makes it very easy to use (`Easiness of use').
	
	\section{Conclusion}
	In this paper we formulated the hierarchical Newton's method of control which enables numerically stable computations in case of singularities of kinematic tasks. We made the link to control and formulated second order PD controllers in the velocity domain with exponential convergence properties even in the case of regularization. Our simulations on toy robotic examples confirmed that such an approach is viable.
	Experiments with the HRP-2Kai, a complex humanoid robot, showed that the reliance on accurate representations of second order information --that is in contrast to approximations like classical damping methods-- leads to better robot workspace occupancy and consequently better task error reduction. At the same time we confirmed that the notion of hierarchy enables formulating problems with strict safety prioritization.
	
	Computation times of the hierarchical least squares solver are a limiting factor. Without important dynamic effects, the active set iteration count is limited to a few iterations. However, situations where several torque, trust region and contact force constraints become active are more challenging for the active set search. Our future work needs to focus on handling these situations cheaply, for example by factorization updates in the solver proposed in~\cite{dimitrov:2015} or a more effective active set search.
	
	\section{Acknowledgment}
	We deeply thank Pierre-Brice Wieber and Dimitar Dimitrov for providing us with the code for LexLSI~\cite{dimitrov:2015} which was indispensable for this work; and acknowledge that Sec.~\ref{sec:minNonLinGeoFct} is the result of fruitful discussions with Pierre-Brice Wieber.
	%%%%%%%%%%%%%%%%%%%%%%%%%%%%%%%%%%%%%%%%%%%%%%%%%%%%%%%%%%%%%%%%%%%%%%%%%%%%%%%%%%%%%%%%%%%%%%%%%%
	%%%%%%%%%%%%%%%%%%%%%%%%%%%%%%%%%%%%%%%%%%%%%%%%%%%%%%%%%%%%%%%%%%%%%%%%%%%%%%%%%%%%%%%%%%%%%%%%%%
	
	\bibliographystyle{IEEEtran}
	\bibliography{bib}

\begin{thebibliography}{10}
\providecommand{\url}[1]{#1}
\csname url@samestyle\endcsname
\providecommand{\newblock}{\relax}
\providecommand{\bibinfo}[2]{#2}
\providecommand{\BIBentrySTDinterwordspacing}{\spaceskip=0pt\relax}
\providecommand{\BIBentryALTinterwordstretchfactor}{4}
\providecommand{\BIBentryALTinterwordspacing}{\spaceskip=\fontdimen2\font plus
\BIBentryALTinterwordstretchfactor\fontdimen3\font minus
  \fontdimen4\font\relax}
\providecommand{\BIBforeignlanguage}[2]{{%
\expandafter\ifx\csname l@#1\endcsname\relax
\typeout{** WARNING: IEEEtran.bst: No hyphenation pattern has been}%
\typeout{** loaded for the language `#1'. Using the pattern for}%
\typeout{** the default language instead.}%
\else
\language=\csname l@#1\endcsname
\fi
#2}}
\providecommand{\BIBdecl}{\relax}
\BIBdecl

\bibitem{Escande2014}
A.~Escande, N.~Mansard, and P.-B. Wieber, ``{Hierarchical quadratic
  programming: Fast online humanoid-robot motion generation},'' \emph{The
  International Journal of Robotics Research}, vol.~33, no.~7, pp. 1006--1028,
  2014.

\bibitem{nakamura1986}
Y.~Nakamura and H.~Hanafusa, ``{Inverse Kinematic Solutions With Singularity
  Robustness for Robot Manipulator Control},'' \emph{J. Dyn. Sys., Meas.,
  Control}, vol. 108, no.~3, pp. 163--171, 1986.

\bibitem{Siciliano1991}
B.~Siciliano and J.-J.~E. Slotine, ``The general framework for managing
  multiple tasks in high redundant robotic systems,'' in \emph{International
  Conference on Advanced Robotics}, 1991, pp. 1211 -- 1216 vol.2.

\bibitem{Chaumette2000}
F.~Chaumette and E.~Marchand, ``A new redundancy-based iterative scheme for
  avoiding joint limits. application to visual servoing,'' in \emph{IEEE
  International Conference on Robotics and Automation}, vol.~2, 2000, pp.
  1720--1725.

\bibitem{Mansard2005}
N.~Mansard and F.~Chaumette, ``A new redundancy formalism for avoidance in
  visual servoing,'' in \emph{IEEE/RSJ International Conference on Intelligent
  Robots and Systems}, 2005, pp. 468--474.

\bibitem{abe:sca:2007}
Y.~Abe, M.~{Da Silva}, and J.~Popovi{\'{c}}, ``{Multiobjective control with
  frictional contacts},'' \emph{ACM SIGGRAPH Symposium on Computer Animation},
  vol.~1, pp. 249--258, 2007.

\bibitem{collette:humanoids:2007}
C.~Collette, A.~Micaelli, C.~Andriot, and P.~Lemerle, ``Dynamic balance control
  of humanoids for multiple grasps and non coplanar frictional contacts,'' in
  \emph{IEEE/RAS International Conference on Humanoid Robots}, Pittsburgh, PA,
  Nov. 29 - Dec. 1 2007, pp. 81--88.

\bibitem{feng:humanoids:2013}
S.~Feng, X.~Xinjilefu, W.~Huang, and C.~G. Atkeson, ``{3D} walking based on
  online optimization,'' in \emph{IEEE-RAS International Conference on Humanoid
  Robots}, 2013.

\bibitem{kuindersma:icra:2014}
S.~Kuindersma, F.~Permenter, and R.~Tedrake, ``An efficiently solvable
  quadratic program for stabilizing dynamic locomotion,'' in \emph{IEEE
  International Conference on Robotics and Automation}, HK, China, 2014.

\bibitem{vaillant:auro:2016}
J.~Vaillant, A.~Kheddar, H.~Audren, F.~Keith, S.~Brossette, A.~Escande,
  K.~Bouyarmane, K.~Kaneko, M.~Morisawa, P.~Gergondet, E.~Yoshida, S.~Kajita,
  and F.~Kanehiro, ``{Multi-contact vertical ladder climbing with an HRP-2
  humanoid},'' \emph{Autonomous Robots}, vol.~40, no.~3, pp. 561--580, 2016.

\bibitem{pfeiffer2017}
K.~{Pfeiffer}, A.~{Escande}, and A.~{Kheddar}, ``Nut fastening with a humanoid
  robot,'' in \emph{IEEE/RSJ International Conference on Intelligent Robots and
  Systems}, Sep. 2017, pp. 6142--6148.

\bibitem{Bouyarmane2019}
K.~Bouyarmane, K.~Chappellet, J.~Vaillant, and A.~Kheddar, ``Quadratic
  programming for multirobot and task-space force control,'' \emph{IEEE
  Transactions on Robotics}, vol.~35, no.~1, pp. 64--77, 2019.

\bibitem{Kanoun2011}
O.~Kanoun, F.~Lamiraux, and P.-B. Wieber, ``{Kinematic control of redundant
  manipulators: generalizing the task priority framework to inequality
  tasks},'' \emph{IEEE Trans. on Robotics}, vol.~27, no.~4, pp. 785--792, 2011.

\bibitem{Herzog2016}
A.~Herzog, N.~Rotella, S.~Mason, F.~Grimminger, S.~Schaal, and L.~Righetti,
  ``Momentum control with hierarchical inverse dynamics on a torque-controlled
  humanoid,'' \emph{Autonomous Robots}, vol.~40, no.~3, pp. 473--491, Mar 2016.

\bibitem{Arechavaleta2017}
G.~Arechavaleta, A.~Morales-Díaz, H.~M. Pérez-Villeda, and M.~Castelán,
  ``Hierarchical task-based control of multirobot systems with terminal
  attractors,'' \emph{IEEE Transactions on Control Systems Technology},
  vol.~25, no.~1, pp. 334--341, 2017.

\bibitem{Yoshikawa1985}
T.~Yoshikawa, ``Manipulability of robotic mechanisms,'' \emph{The International
  Journal of Robotics Research}, vol.~4, no.~2, pp. 3--9, 1985.

\bibitem{Johansen2004}
T.~Johansen, T.~Fossen, and S.~Berge, ``Constrained nonlinear control
  allocation with singularity avoidance using sequential quadratic
  programming,'' \emph{IEEE Transactions on Control Systems Technology},
  vol.~12, no.~1, pp. 211--216, 2004.

\bibitem{Leeghim2009}
H.~Leeghim, I.-H. Lee, D.-H. Lee, H.~Bang, and J.-O. Park, ``Singularity
  avoidance of control moment gyros by predicted singularity robustness: Ground
  experiment,'' \emph{IEEE Transactions on Control Systems Technology},
  vol.~17, no.~4, pp. 884--891, 2009.

\bibitem{Maciejewski1988}
A.~A. Maciejewski and C.~A. Klein, ``{Numerical filtering for the operation of
  robotic manipulators through kinematically singular configurations},''
  \emph{Journal of Robotic Systems}, vol.~5, no.~6, pp. 527--552, 1988.

\bibitem{Khalil2002}
W.~Khalil and E.~Dombre, ``Chapter 5- direct kinematic model of serial
  robots,'' in \emph{Modeling, Identification and Control of Robots}, W.~Khalil
  and E.~Dombre, Eds.\hskip 1em plus 0.5em minus 0.4em\relax
  Butterworth-Heinemann, 2002, pp. 85 -- 115.

\bibitem{Bianco2020}
C.~Guarino Lo~Bianco and M.~Raineri, ``An experimentally validated technique
  for the real-time management of wrist singularities in nonredundant
  anthropomorphic manipulators,'' \emph{IEEE Transactions on Control Systems
  Technology}, vol.~28, no.~4, pp. 1611--1620, 2020.

\bibitem{Chiaverini1997}
S.~Chiaverini, ``{Singularity-robust task-priority redundancy resolution for
  real-time kinematic control of robot manipulators},'' \emph{IEEE Transactions
  on Robotics and Automation}, vol.~13, no.~3, pp. 398--410, 1997.

\bibitem{kheddar2019ram}
A.~Kheddar, S.~Caron, P.~Gergondet, A.~Comport, A.~Tanguy, C.~Ott, B.~Henze,
  G.~Mesesan, J.~Englsberger, M.~A. Roa, P.-B. Wieber, F.~Chaumette,
  F.~Spindler, G.~Oriolo, L.~Lanari, A.~Escande, K.~Chappellet, F.~Kanehiro,
  and P.~Rabaté, ``Humanoid robots in aircraft manufacturing: The airbus use
  cases,'' \emph{IEEE Robotics Automation Magazine}, vol.~26, no.~4, pp.
  30--45, 2019.

\bibitem{kumagai2019ram}
I.~Kumagai, M.~Morisawa, T.~Sakaguchi, S.~Nakaoka, K.~Kaneko, H.~Kaminaga,
  S.~Kajita, M.~Benallegue, R.~Cisneros, and F.~Kanehiro, ``Toward
  industrialization of humanoid robots: Autonomous plasterboard installation to
  improve safety and efficiency,'' \emph{IEEE Robotics Automation Magazine},
  vol.~26, no.~4, pp. 20--29, 2019.

\bibitem{Nishiwaki2009}
K.~Nishiwaki and S.~Kagami, ``Online walking control system for humanoids with
  short cycle pattern generation,'' \emph{The International Journal of Robotics
  Research}, vol.~28, no.~6, pp. 729--742, 2009.

\bibitem{Mason2016}
S.~{Mason}, N.~{Rotella}, S.~{Schaal}, and L.~{Righetti}, ``Balancing and
  walking using full dynamics {LQR} control with contact constraints,'' in
  \emph{IEEE-RAS International Conference on Humanoid Robots}, Nov 2016, pp.
  63--68.

\bibitem{bonnet:hal-02048085}
V.~Bonnet, K.~Pfeiffer, P.~Fraisse, A.~Crosnier, and G.~Venture,
  ``Self-generation of optimal exciting motions for identification of a
  humanoid robot,'' \emph{International Journal of Humanoid Robotics}, vol.~15,
  no.~6, p. 1850024, Dec. 2018.

\bibitem{ChiaveriniSiciliano1994}
S.~{Chiaverini}, B.~{Siciliano}, and O.~{Egeland}, ``Review of the damped
  least-squares inverse kinematics with experiments on an industrial robot
  manipulator,'' \emph{IEEE Transactions on Control Systems Technology},
  vol.~2, no.~2, pp. 123--134, June 1994.

\bibitem{Buss2005}
S.~R. Buss and J.-S. Kim, ``Selectively damped least squares for inverse
  kinematics,'' \emph{Journal of Graphics Tools}, vol.~10, no.~3, pp. 37--49,
  2005.

\bibitem{Sugihara2011}
T.~Sugihara, ``Solvability-unconcerned inverse kinematics by the
  levenberg–marquardt method,'' \emph{IEEE Transactions on Robotics},
  vol.~27, no.~5, pp. 984--991, October 2011.

\bibitem{Harish2016}
P.~Harish, M.~Mahmudi, B.~L. Callennec, and R.~Boulic, ``Parallel inverse
  kinematics for multithreaded architectures,'' \emph{ACM Transactions on
  Graphics}, vol.~35, no.~2, pp. 1--13, Feb. 2016.

\bibitem{Dennis1981}
J.~E. Dennis, Jr., D.~M. Gay, and R.~E. Walsh, ``An adaptive nonlinear
  least-squares algorithm,'' \emph{ACM Trans. Math. Softw.}, vol.~7, no.~3, pp.
  348--368, Sep. 1981.

\bibitem{Deo:1993}
A.~S. Deo and I.~D. Walker, ``Adaptive non-linear least squares for inverse
  kinematics,'' in \emph{IEEE International Conference on Robotics and
  Automation}, vol.~1, May 1993, pp. 186--193.

\bibitem{Pfeiffer2018}
K.~{Pfeiffer}, A.~{Escande}, and A.~{Kheddar}, ``Singularity resolution in
  equality and inequality constrained hierarchical task-space control by
  adaptive nonlinear least squares,'' \emph{IEEE Robotics and Automation
  Letters}, vol.~3, no.~4, pp. 3630--3637, Oct 2018.

\bibitem{Broyden1970}
C.~G. Broyden, ``{The Convergence of a Class of Double-rank Minization
  Algorithms},'' \emph{Journal of the Mathematics and its Applications},
  vol.~6, pp. 76--90, 1970.

\bibitem{Nenchev2000}
D.~N. Nenchev, Y.~Tsumaki, and M.~Uchiyama, ``Singularity-consistent
  parameterization of robot motion and control,'' \emph{The International
  Journal of Robotics Research}, vol.~19, no.~2, pp. 159--182, 2000.

\bibitem{Wang2010}
J.~Wang, Y.~Li, and X.~Zhao, ``Inverse kinematics and control of a 7-dof
  redundant manipulator based on the closed-loop algorithm,'' \emph{Int.
  Journal of Advanced Robotic Systems}, vol.~7, no.~4, p.~37, 2010.

\bibitem{Siciliano2007}
B.~Siciliano and O.~Khatib, \emph{Springer Handbook of Robotics}.\hskip 1em
  plus 0.5em minus 0.4em\relax Berlin, Heidelberg: Springer-Verlag, 2007.

\bibitem{dimitrov:2015}
\BIBentryALTinterwordspacing
D.~Dimitrov, A.~Sherikov, and P.-B. Wieber, ``{Efficient resolution of
  potentially conflicting linear constraints in robotics},'' Aug. 2015, v1.
  [Online]. Available: \url{https://hal.inria.fr/hal-01183003}
\BIBentrySTDinterwordspacing

\bibitem{Yenamandra2019}
T.~Yenamandra, F.~Bernard, J.~Wang, F.~Mueller, and C.~Theobalt, ``Convex
  optimisation for inverse kinematics,'' in \emph{International Conference on
  3D Vision}, Los Alamitos, CA, USA, sep 2019, pp. 318--327.

\bibitem{Dai2019}
H.~Dai, G.~Izatt, and R.~Tedrake, ``Global inverse kinematics via mixed-integer
  convex optimization,'' \emph{The International Journal of Robotics Research},
  vol.~38, no. 12-13, pp. 1420--1441, 2019.

\bibitem{DeLasa2010}
M.~de~Lasa, I.~Mordatch, and A.~Hertzmann, ``{Feature-based locomotion
  controllers},'' \emph{ACM Transactions on Graphics}, vol.~29, no.~4, p.~1,
  2010.

\bibitem{Saab2013}
L.~Saab, O.~E. Ramos, F.~Keith, N.~Mansard, P.~Soueres, and J.~Y. Fourquet,
  ``{Dynamic whole-body motion generation under rigid contacts and other
  unilateral constraints},'' \emph{IEEE Transactions on Robotics}, vol.~29,
  no.~2, pp. 346--362, 2013.

\bibitem{Erleben2017}
K.~Erleben and S.~Andrews, ``Inverse kinematics problems with exact hessian
  matrices,'' in \emph{International Conference on Motion in Games}, 2017, pp.
  14:1--14:6.

\bibitem{Golub1996}
G.~H. Golub and C.~F. Van~Loan, \emph{Matrix Computations (3rd Ed.)}.\hskip 1em
  plus 0.5em minus 0.4em\relax Baltimore, MD, USA: Johns Hopkins University
  Press, 1996.

\bibitem{Higham1986}
N.~Higham, ``Computing the polar decomposition with applications,'' \emph{SIAM
  Journal on Scientific and Statistical Computing}, vol.~7, no.~4, pp.
  1160--1174, 1986.

\bibitem{Udwadia1992}
F.~Udwadia and R.~E.~Kalaba, ``A new perspective on constrained motion,''
  \emph{Proceedings of The Royal Society A: Mathematical, Physical and
  Engineering Sciences}, vol. 439, pp. 407--410, 11 1992.

\bibitem{Udwadia2010}
F.~Udwadia and A.~Schutte, ``Equations of motion for general constrained
  systems in lagrangian mechanics,'' \emph{Acta Mech}, vol. 213, 08 2010.

\bibitem{gill:techrep:1986}
P.~E.~E. Gill, S.~J. Hammarling, W.~Murray, M.~A. Saunders, and M.~H. Wright,
  ``User's guide for lssol (v~1.0): a fortran package for constrained linear
  least-squares and convex quadratic programming,'' Stanford University, Tech.
  Rep. 86-1, January 1986.

\bibitem{faverjon1987}
\BIBentryALTinterwordspacing
B.~Faverjon and P.~Tournassoud, ``{A local based approach for path planning of
  manipulators with a high number of degrees of freedom},'' {INRIA}, Tech. Rep.
  RR-0621, Feb. 1987. [Online]. Available:
  \url{https://hal.inria.fr/inria-00075933}
\BIBentrySTDinterwordspacing

\bibitem{nitsche:iros:2015}
M.~Nitsche, T.~Krajn{\'\i}k, P.~{\v C}{\' i}{\v z}ek, M.~Mejail, and
  T.~Duckett, ``{WhyCon}: an efficient, marker-based localization system,'' in
  \emph{IEEE/RAS IROS Workshop on Open Source Aerial Robotics}, 2015.

\bibitem{pham2013}
Q.~C. Pham, ``A general, fast, and robust implementation of the time-optimal
  path parameterization algorithm,'' \emph{IEEE Transactions on Robotics},
  vol.~30, pp. 1533--1540, 2013.

\end{thebibliography}
	
	\balance

\end{document}